\documentclass[pmlr,twocolumn,10pt]{jmlr} %

\usepackage{booktabs}
\usepackage{lmodern}
\usepackage{times}
\usepackage{siunitx}

\usepackage[switch]{lineno}
\usepackage{amsmath}
\usepackage{amssymb}
\usepackage{graphicx}
\usepackage{mathrsfs}
\usepackage{amsfonts}
\usepackage{booktabs} %
\usepackage{caption}  %
\usepackage{threeparttable} %
\usepackage{algorithm}
\usepackage{algorithmicx}
\usepackage{algpseudocode}
\usepackage{listings}
\usepackage{enumitem}
\usepackage{chngcntr}

\usepackage{booktabs}

\usepackage{lipsum}
\usepackage{subcaption}

\usepackage[T1]{fontenc}    %
\usepackage{csquotes}       %
\usepackage{diagbox}
\usepackage{comment}
\usepackage{mathtools}
\usepackage{makecell}  %
\usepackage{tcolorbox}
\usepackage{bbm}

\DeclareMathOperator*{\argmax}{arg\,max}

\newcommand{\division}{\mkern-\medmuskip\rotatebox[origin=c]{45}{\scalebox{0.9}{$-$}}\mkern-\medmuskip}

\theorembodyfont{\upshape}
\theoremheaderfont{\scshape}
\theorempostheader{:}
\theoremsep{\newline}

\jmlrpages{}
\title[Aim-Limit-Defined Multi-Objective System For Cervical Cancer Brachytherapy Treatment Planning]{ALMo: Interactive Aim-Limit-Defined, Multi-Objective System for Personalized High-Dose-Rate Brachytherapy Treatment Planning and Visualization for Cervical Cancer}

\author{%
 \Name{Edward Chen} \Email{edjchen@stanford.edu}\\
 \Name{Natalie Dullerud} \Email{ndulleru@stanford.edu}\\
 \addr Department of Computer Science, Stanford University\\
 \Name{Pang Wei Koh} \Email{pangwei@cs.washington.edu}\\
 \addr Paul G. Allen School of Computer Science \& Engineering, University of Washington\\
 \Name{Thomas Niedermayr} \Email{trn@stanford.edu}\\
 \Name{Elizabeth Kidd} \Email{ekidd@stanford.edu}\\
 \addr Department of Radiation Oncology, Stanford University School of Medicine\\
 \Name{Sanmi Koyejo} \Email{sanmi@stanford.edu}\\
 \Name{Carlos Guestrin} \Email{guestrin@stanford.edu}\\
 \addr Department of Computer Science, Stanford University
}

\begin{document}

\maketitle

\begin{abstract}
    In complex clinical decision-making, clinicians must often track a variety of competing metrics defined by ``aim'' (ideal) and ``limit'' (strict) thresholds. Sifting through these high-dimensional tradeoffs to infer the optimal patient-specific strategy is cognitively demanding and historically prone to variability. In this paper, we address this challenge within the context of High-Dose-Rate (HDR) brachytherapy for cervical cancer, where planning requires strictly managing radiation hot spots while balancing tumor coverage against organ sparing. We present ALMo (Aim-Limit-defined Multi-Objective system), an interactive decision support system designed to infer and operationalize clinician intent. ALMo employs a novel optimization framework that minimizes manual input through automated parameter setup and enables flexible control over toxicity risks. \textit{Crucially, the system allows clinicians to navigate the Pareto surface of dosimetric tradeoffs by directly manipulating intuitive aim and limit values.} In a retrospective evaluation of 25 clinical cases, ALMo generated treatment plans that consistently met or exceeded manual planning quality, with 65\% of cases demonstrating dosimetric improvements. Furthermore, the system significantly enhanced efficiency, reducing average planning time to approximately 17 minutes, compared to the conventional 30--60 minutes. While validated in brachytherapy, ALMo demonstrates a generalized framework for streamlining interaction in multi-criteria clinical decision-making.
\end{abstract}

\section{Introduction}
\label{sec:intro}
High-dose-rate (HDR) brachytherapy for cervical cancer operates in a high-stakes clinical environment where sealed radiation sources are positioned directly within or adjacent to cancerous tissue. The treatment planning process presents a formidable optimization challenge because it must be performed intraoperatively while the patient remains under anesthesia with applicators in place. Achieving an optimal treatment plan is essential for clinical success, as it directly dictates tumor control and the severity of acute or long-term toxicity. However, significant anatomical variability among patients creates an expansive solution space that complicates the search for ideal dosimetric distributions. This time-sensitive nature of intraoperative planning is particularly critical, as extended planning durations correlate with prolonged anesthesia exposure, increased patient discomfort, and elevated risks of positional uncertainties that may compromise treatment accuracy \citep{craft_improved_2012}.

To address these complexities, multi-objective optimization (MOO) techniques have emerged as tools for HDR brachytherapy planning \citep{ruotsalainen_interactive_2010, deufel_pnav_2020, dinkla_comparison_2015}. These methods assist clinicians in navigating the inherent trade-offs between competing objectives, primarily the maximization of tumor coverage versus the minimization of radiation exposure to organs-at-risk (OARs). The central tension typically involves ensuring sufficient radiation to the planning target volume (PTV) while protecting adjacent healthy tissues such as the bladder, rectum, and bowel. For example, increasing tumor coverage from 90\% to 95\% may theoretically improve efficacy, but such an adjustment often risks exceeding safe radiation thresholds for nearby organs. While MOO frameworks provide a mathematical basis for these decisions, they often struggle to fully capture the nuanced preferences required in clinical practice.

This gap between algorithmic potential and practical utility manifests as significant limitations regarding plan quality, computational efficiency, and user interaction. From a quality standpoint, existing optimizers frequently fail to explicitly account for radiation toxicity hot spots, defined as regions receiving 200\% of the prescription dosage \citep{moren_mathematical_2019}, in the sensitive areas surrounding the PTV and OARs. Efficiency is also a major concern, as the high-dimensional nature of the dosage trade-off space typically requires extensive computational time that hinders rapid, patient-specific exploration. Furthermore, current systems lack interactive mechanisms to incorporate institutional protocols directly, such as American Brachytherapy Society guidelines that specify ideal "aim" values (e.g., 513 cGy for the bladder) and strict "limit" constraints (e.g., 601 cGy) \citep{romano_high_2018}.

In this paper, we present ALMo (Aim-Limit-defined Multi-Objective system), an interactive decision support system designed to bridge the gap between algorithmic optimization and clinical workflow. ALMo operationalizes clinical intuition by allowing users to directly manipulate aim and limit values to navigate the dosage trade-off surface. Unlike existing tools that often require substantial manual setup or ignore critical toxicity metrics \citep{deufel_pnav_2020}, ALMo employs a three-stage pipeline. This system automates initialization, enables direct control over radiation hot spots in sensitive regions \citep{carrara_comparison_2017, limkin_vaginal_2016}, and provides an efficient mechanism for iterative plan refinement, addressing a common usability barrier in MOO frameworks \citep{kyroudi_discrepancies_2016, bokrantz_projections_2015}. Through this approach, ALMo streamlines the planning process to simultaneously enhance \textbf{quality}, \textbf{efficiency}, and \textbf{interaction}.

We retrospectively evaluated ALMo on 25 clinical cases to demonstrate its practical utility and performance. Regarding plan quality, the system consistently generated clinically viable plans, with 65\% of cases showing minor to major improvements in dosimetric metrics compared to manual planning. In terms of efficiency, ALMo reduced the average planning time to approximately 17.6 minutes, an improvement over traditional workflows that typically range from 30 to 90 minutes. Finally, our evaluation of clinician interaction revealed that the system's integrated visualization tools were sufficient for plan assessment in 66.7\% of cases, significantly reducing the reliance on external software and confirming the system's ability to facilitate effective clinical decision-making.

\section{Related Works}\label{sec:related-works}

MOO has emerged as a prominent methodology for enhancing the efficiency and quality of HDR brachytherapy treatment planning. Several studies have demonstrated that MOO frameworks can streamline the planning process by automating the trade-off analysis between competing dosimetric objectives \citep{kierkels_multicriteria_2015, kamran_multi-criteria_2016, xiao_multi-criteria_2018}. While approaches such as those by \citet{cui_multi-criteria_2018} and \citet{cui_multi-criteria_2018-1} allow clinicians to review batches of proposed plans, these systems often restrict users to a single iteration or lack the mechanisms for fine-grained exploration of specific metric constraints \citep{dickhoff_versatility_2025, goli_small_2018, oud_fast_2020}. Similarly, while \citet{belanger_gpu-based_2019} focused on accelerating these optimizations via GPU-based architectures, hardware acceleration alone does not resolve the fundamental challenges regarding clinical usability and preference elicitation.

Despite current advancements, existing interactive tools face significant limitations in their workflow integration and dosimetric scope. For instance, the Pareto frontier navigation tool (PNaV) introduced by \citet{deufel_pnav_2020} enables exploration across a 30\% dose range but requires substantial manual parameter setup and notably excludes critical hot spot metrics. In a different vein, \citet{jafarzadeh_penalty_2024} proposed a multi-objective Bayesian optimization framework to determine optimal penalty weights. However, their approach prioritizes the exhaustive population of the Pareto frontier rather than facilitating a clinically guided exploration of the solution space. Consequently, these methods often lack efficient mechanisms for iterative plan refinement that allow clinicians to navigate within relevant clinical bounds \citep{kyroudi_discrepancies_2016, bokrantz_projections_2015, belanger_commissioning_2022}.

Beyond interactive navigation, a parallel research thread has explored automated preference learning to reduce clinician burden. Inverse optimization methods have been developed to infer objective function weights from historical treatment plans, enabling prediction of clinician-acceptable parameters for new patients \citep{babier_inverse_2018, boutilier_models_2015}. More recently, \citet{ajayi_objective_2022} proposed a bilevel optimization framework for sparse objective selection in prostate cancer treatment. While these data-driven approaches reduce planning time, they typically require large training datasets and may not generalize across institutions with differing protocols. 

Deep learning approaches have also emerged for dose prediction in cervical brachytherapy \citep{ma_dose_2022, yu_cnn-based_2024, gronberg_technical_2021, liu_technical_2021, babier_openkbp_2021}, though these methods focus on predicting achievable dose distributions rather than facilitating interactive trade-off exploration. ALMo distinguishes itself by combining the safety benefits of constrained optimization with an interactive framework that preserves clinician agency through intuitive aim-limit manipulation. We further stress the added controllability of our approach as a major advantage over deep learning ones.

A particularly critical gap in current MOO frameworks is the inadequate management of radiation toxicity in sensitive anatomical regions. Neglecting control over hot spots in the cervical and vaginal mucosa can significantly impact patient morbidity \citep{carrara_comparison_2017, limkin_vaginal_2016}. Integrating these additional constraints into standard optimization models typically results in a combinatorial explosion of weight parameters, rendering the problem computationally intractable for real-time clinical use. Without integrating explicit constraints, \citet{carrara_comparison_2017} compares various different optimization algorithms' implicit effectiveness towards reducing toxicity levels. \citet{moren_technical_2023} proposes to include a general proximity-based penalty to the optimization model. ALMo addresses these deficiencies by providing a system that integrates automated parameter setup, \textit{explicit} toxicity control for sensitive tissues, and an interactive aim-limit interface for intuitive Pareto surface navigation.

\section{Background}
\label{sec:background}

We follow a similar setup and notation as described in \citet{deufel_pnav_2020}. For each patient, we are given a set of computed tomography (CT) slices, where the planning target volume (PTV) and organs-at-risk (OAR) (bladder, rectum, and bowel) contours have been pre-segmented. We denote the set of voxels as $V$ and the set of OAR as $O$. The set of applicators we use for our experiments consists of the tandem, right and left ovoids, and a variable number of needles. We are provided the dwell positions of the applicators, denoted as $T$. Let $d \in \mathbb{R}^{|V|}$ be the radiation dosage distribution, $t \in \mathbb{R}^{|T|}$ be the dwell times for each of the dwell locations, and $G \in \mathbb{R}^{|V| \times |T|}$ be the radiation dosage rate matrix, where $G_{ij}$ represents the radiation dosage (in units of cGY) deposited in voxel $i$ from the dwell source $j$ at unit strength ($t=1$ second). This was calculated with the TG-43 dose calculation formalism \citep{rivard_update_2004}. Thus, the above values are related by $d = Gt$.

The clinical objectives are represented by a vector-valued function $f = (f_1,...,f_{|K|}): \mathbb{R}_+^{|V|} \rightarrow \mathbb{R}_+^{|K|}$ of $d \in D$, where each component $f_k$ is a dosimetric metric for a structure $k$ in the set of optimized structures $K$. These are standard dose-volume-histogram (DVH) metrics, such as $\text{PTV}_{V700}$ or Bladder $D_{2cc}$. Since standard DVH metrics are often non-convex, we use a convex approximation for each objective known as the Truncated Conditional Value-at-Risk (TCVaR) metric \citep{wu_new_2020}. TCVaR can be applied to either the high-dose or low-dose tail of a distribution. For OARs, where the goal is to limit high doses, we apply the metric to the high-dose tail and denote it as $\theta_k^+$. Conversely, for the PTV, where the goal is to maximize coverage, we apply it to the low-dose tail and denote it as $\theta_k^-$. Additional mathematical description and background information on multi-objective optimization are provided in Appendices \ref{app:background-metrics} and \ref{sec:moo-background}.

\section{ALMo: Decision Support System}
\label{sec:almo-system}

\begin{figure}
    \centering
    \includegraphics[width=0.5\textwidth]{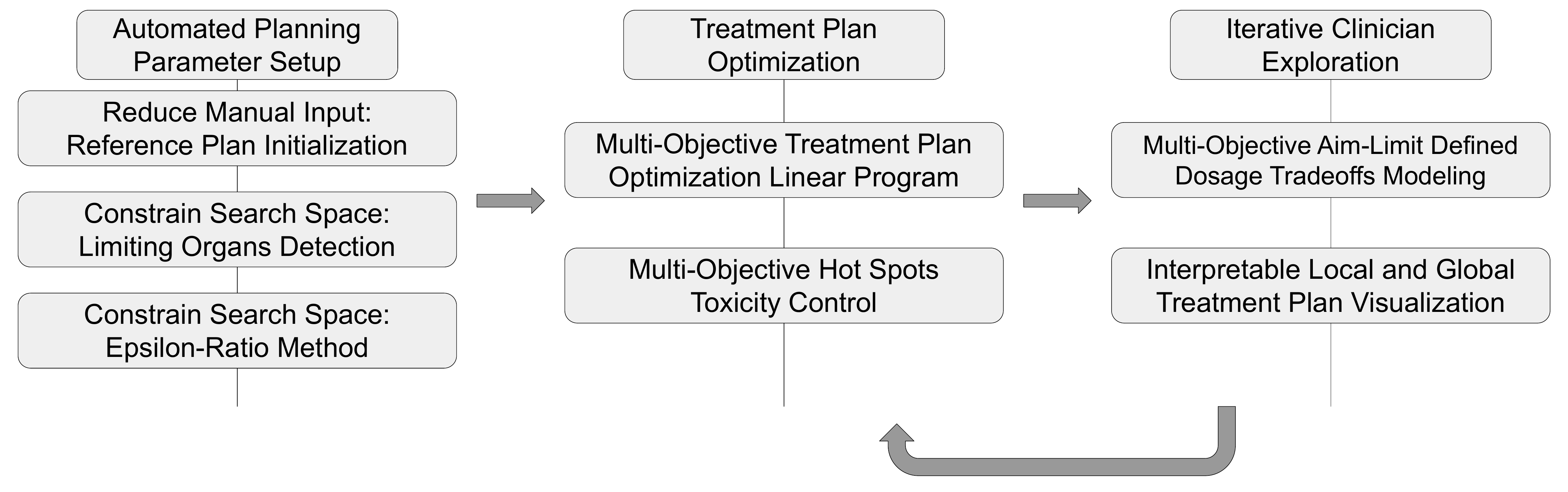}
    \caption{Diagram of the full pipeline for ALMo. Automated Planning Parameter Setup only happens once in the beginning of the pipeline, whereas Treatment Plan Optimization and Iterative Clinician Exploration both contribute to the interactive nature of the tool. 
    }
    \label{fig:full-pipeline}
\end{figure}

ALMo consists of three components designed to improve the quality, efficiency, and interaction capabilities of the cervical cancer brachytherapy treatment planning process: (1) Treatment Plan Optimization, which uses a multi-objective optimization linear program to resolve a single treatment plan on the Pareto frontier, while controlling for hot spots toxicity values, (2) Iterative Clinician Exploration, which assists the clinician with navigating across subsets of the high-dimensional Pareto frontier by way of multi-objective aim and limit slider bars and interpretable visualization methods, and (3) Automated Planning Parameter Setup, which comprises of a series of algorithms designed to improve the efficiency of the tool by pre-calculating starting points with minimal manual input. A figure of the entire pipeline is shown in Figure \ref{fig:full-pipeline}. Within ALMo, Automated Planning Parameter Setup occurs first, sequentially, however we organize our sections as ordered above for a more intuitive explanation.

\subsection{Treatment Plan Optimization}\label{inner-loop-section}

We lay out the foundation of our treatment plan optimization in Section \ref{mco-lp-section} and use it to control for toxicity in Section \ref{hot-spots-section}.

\subsubsection{Multi-Objective Plan Optimization}\label{mco-lp-section}
To construct a library of HDR treatment plans representing different clinical trade-offs, we use an $\epsilon$-constraint multi-objective optimization approach \citep{romeijn_novel_2003, holm_linear_2013, deufel_pnav_2020}. This method generates Pareto-optimal plans by solving a sequence of convex optimization problems. Each problem maximizes PTV coverage while ensuring that OAR doses do not exceed specific limits, $\epsilon_k$. The general form of this problem is:
\begin{equation}
\begin{array}{ll@{}ll}
\text{maximize}  & \theta_1^-(\cdot) \\
\text{subject to}& \theta_k^+(\cdot) \leq \epsilon_k, & \hspace{1cm} k=2,...,|K| \\
& d = Gt \\
& t \geq 0
\end{array}
\label{equation:moo}
\end{equation}
where $k=1$ corresponds to the PTV structure. Constructing the finite library of treatment plans is formulated as a sequence of linear programs, (Equation \ref{equation:moo}), centered around a reference treatment plan with the CVaR metrics $\gamma^{+,init} = (\gamma_1^{+,init},\dots,\gamma_{|K|}^{+,init})$. The reference treatment plan is based on an initial dosage distribution $d^{init}$ ($\Bar{d}=d^{init}$). To specify the trade-off library resolution, we use a set of metric value ratios, denoted as $R_k$ for structure $k$. For instance, to obtain a desired range of the kth metric as approximately $\pm$ 15\% of $f_k(d^{init})$, with resolution ($\Delta_k$) of 10\%, we use $r_k \in R_k = \{-0.15, -0.05, 0.05, 1.15\}$. Formally, 
\begin{equation}\label{eq:r_set_def}
    R_k = \{r_k \mid r_k = \ell_k + c\Delta_k, c \in \mathbb{N}, r_k \leq u_k\}    
\end{equation}
where $\ell_k$, $u_k$ denote the lower and upper bounds, respectively. Each of the values $r_k$ $\in$ $R_k$ is multiplied by $\gamma^{+,init}_k$ to determine the values $\epsilon_k$ on the right-hand sides of the first set of constraints in the linear program above. We denote optimization problem (Equation \ref{equation:moo}) as $g: R \subseteq \mathbb{R}_+^{|K|-1} \rightarrow D \subseteq \mathbb{R}_+^{|V|}$. For notation purposes, we use $r \in R$ to denote an individual treatment plan. To obtain a representation of the Pareto frontier, we perform a grid search over the set $R = R_2 \bigtimes ... \bigtimes R_{|K\backslash \{PTV\}|}$ (where $K$ is the set of structures we optimize over), which depending on $|R_k|$, can become extremely computationally demanding due to the combinatorial explosion of values. Section \ref{preprocessing-section} discusses how we propose to resolve that.

\subsubsection{Multi-Objective Hot Spots Toxicity Control}\label{hot-spots-section}
In a related study \citep{van_der_meer_better_2018}, it was found that not controlling for hot spots in the areas surrounding the PTV resulted in undesirable dosage distributions due to radiation toxicity levels. To address such an issue, we use a method, similar to \citet{moren_optimization_2021}, in which artificial structures are added to the optimization model (\ref{equation:moo}) to obtain finer-grained control over the dosage distribution in those regions. The artificial structures target the following hot spot regions (HSRs) - cervical, vaginal mucosa, and surrounding needle regions \citep{ladbury_practical_2023, elburg_assessment_2023}. We denote the set of HSRs as $H$, where $H \cup O \cup \{PTV\} = K$ and $O$ is the set of OAR.

The following steps are done to create the artificial structure for the cervical and vaginal mucosa regions: (1) create an artificial ring which consists of all voxels within a distance of $\psi_{cm}$ from the PTV contour, (2) set only the dwell times for the ovoids, which are initialized from the standard point A normalization loading pattern, also described by \citet{kirisits_dose_2005}, (3) select the hottest $\omega\%$ of the voxels in the artificial ring, from the resulting dosage distribution from (2), to be used as the mucosa region artificial structure, (4) use remaining voxels in the artificial ring as the cervical region artificial structure. The intuition behind only setting the dwell times for the ovoids is the ovoid applicators are often positionally set to be closest to the vaginal mucosa region \citep{serban_vaginal_2021}. The artificial structure surrounding the needle is then defined as all voxels within a distance of $\psi_{n}$ from each of the needle dwell positions. Exact parameter values are in Appendix \ref{sec:experiental-setup}.

\subsection{Iterative Clinician Exploration}\label{outer-loop-section}

\subsubsection{Multi-Objective Aim-Limit Defined Dosage Trade-Offs Modeling}\label{mc-modeling-section}

As validation of each treatment plan displayed to the clinician is time-consuming, we aim to display only a sparse set of treatment plans, $M$, from a subset of the tradeoff surface defined by the clinician's aim and limits, denoted as $\alpha_k^{aim}$ and $\alpha_k^{lim}$, respectively, for each $f_k$ $\forall k \in K$. Consider one of the objectives described earlier in Section \ref{sec:background} -- $PTV_{V700}$. For a given patient, the clinician will often wish to administer a treatment plan which achieves a minimum desirable level, say 90\% $\text{PTV}_{V700}$, in order to decrease the tumor size. In this case, a higher $\text{PTV}_{V700}$ value, say 93\%, would be even better. However, upon reaching a certain threshold, typically 95\%, additional gains in $\text{PTV}_{V700}$ typically offer diminishing utility due to various clinical reasons, such as radiation toxicity and other side effects (i.e. decreasing utility in the other objectives), and eventually saturate. In this case, the utility may saturate at 100\% $\text{PTV}_{V700}$, which we term as the saturation point -- additional gains provide no utility.

\noindent
\textbf{Operationalizing Clinical Priors.}
To operationalize these clinical priors, we translate the aim and limit values into a mathematical utility function, based on the concept of soft-hard utility functions (SHFs) \citep{chen_mosh_2024}. This function formalizes the notion of soft and hard bounds: it heavily penalizes plans that violate the limits, rewards improvements towards the aim values, and reflects the diminishing returns for gains beyond the aim. Specifically, assuming that we want to maximize $f_k$, SHFs maps $f_k$ to a utility space in which (1) the values of $f_k$ which fall below $\alpha_k^{lim}$ map to $-\inf$, the mapping is (2) concave when $f_k \geq \alpha_k^{aim}$, (3) saturated when $f_k \geq \alpha_k^\tau$, and is (4) overall monotonically increasing in $f_k$. %

Following \citet{chen_mosh_2024}, we define the SHF utility function to be as follows:
\begin{equation}
\small %
    u_{\varphi}(x) = \begin{cases}
        1 + 2 \beta (\tilde{\alpha}^\tau - \tilde{\alpha}^{aim}) & \varphi(x) \geq \alpha^{\tau} \\
        1 + 2 \beta (\tilde{\varphi}(x) - \tilde{\alpha}^{aim}) & \alpha^{aim} < \varphi(x) < \alpha^{\tau} \\
        1 & \varphi(x) = \alpha^{aim} \\
        2 \tilde{\varphi}(x) & \alpha^{lim} < \varphi(x) < \alpha^{aim} \\
        0 & \varphi(x) = \alpha^{lim} \\
        -\infty & \varphi(x) < \alpha^{lim} \\
    \end{cases}
\end{equation}
where $\tilde{\varphi}(x)$ and $\tilde{\alpha}$ are the soft (aim) -hard (limit) bound normalized values\footnote{Normalization, for value $z$, is performed according to the aim and limit values, $\alpha^{aim}$ and $\alpha^{lim}$, respectively, using: $\tilde{z} = ((z-\alpha^{lim}) \division (\alpha^{aim}-\alpha^{lim})) * 0.5$}, $\alpha^{\tau}$, the saturation point, determines where the utility values begin to saturate \footnote{We determine $\alpha^{\tau}$ to be $\alpha^{lim}$+$\zeta(\alpha^{aim}-\alpha^{lim})$, for $\zeta > 1.0$. We set $\zeta=2.0$.}, and $\beta$ $\in [0, 1]$ determines the fraction of the original rate of utility, in $[\alpha^{lim},\alpha^{aim}]$, obtained within $[\alpha^{aim}, \alpha^{\tau}]$. In ALMo, we find that SHFs naturally encode the clinician-defined constraints, represented by $\{\alpha^{aim},\alpha^{lim}\}$, allowing us to leverage SHFs to obtain treatment plans which respect such bounds with the following optimization problem $\max_{r \in R} s_{\pmb{\lambda}}(u_{f}(g(r)))$ where $u_f := [u_{f_1}, \dots, u_{f_{|K|}}]$. 

\noindent
\textbf{Obtaining Sparse Set Of Treatment Plans.}
Given that the clinician's precise preferences are unknown, our objective is not to find a single optimal plan. Instead, we aim to present a small, diverse set of high-quality plans that is robustly representative of the solutions within the user-defined bounds. This set is intentionally kept small to avoid overwhelming the clinician, thereby reducing cognitive load. 

Formally, we assume the clinician has a hidden set of preferences, $\pmb{\lambda}^*$, among the $K$ DVH metrics, such that the solution to $\max_{r \in R} s_{\pmb{\lambda}^*} (u_f(g(r)))$ represents the ideal solution on the dosage trade-off surface. Therefore, we wish to return the set of treatment plans, $M$, which contains the unknown $m^* = \argmax_{r \in R} s_{\pmb{\lambda}^*} (u_f(g(r)))$, i.e. the ideal treatment plan that the clinician would like for the patient. Since $\pmb{\lambda}^*$ is known to us, we want a set $M$ which is robust against all potential preferences among the $K$ DVH metrics. Furthermore, $|M| \leq \varrho$ for some small integer value $\varrho$ as we want to avoid overwhelming the clinician with too many treatment plan options. As a result, we formulate our general problem of obtaining the sparse set of treatment plans $M$ from the regions defined by the soft and hard bounds as:
\begin{equation}
    \label{eq:submodular-formulation}
    \max_{M \subseteq R, |M| \leq \varrho} \min_{\pmb{\lambda} \in \Lambda}\Bigg[\dfrac{\max_{r \in M} s_{\pmb{\lambda}}(u_f(g(r)))}{\max_{r \in R} s_{\pmb{\lambda}} (u_f(g(r)))}\Bigg]
\end{equation}
\textit{While \citet{chen_mosh_2024} proposed a general two-step framework, MoSH, for solving Equation \ref{eq:submodular-formulation}, we conceptualize the specific application of it to brachytherapy next}. Step 1, MoSH-Dense, aims to obtain a dense set of Pareto optimal treatment plans according to the aim and limits, which step 2, MoSH-Sparse, then sparsifies before returning to the clinician.

\noindent
\textbf{Step 1: MoSH-Dense.}
The goal for this step is to obtain a dense set of Pareto optimal treatment plans, which are robust against the worst-case potential $\pmb{\lambda^*}$, according to the aim and limit values set by the clinician. As a result, the problem formulation is similar to Equation \ref{eq:submodular-formulation}, however, for computational feasibility, we use the average-case maximization. As a result, we formulate the problem of sampling treatment plans within the aim-limit regions as:
\begin{equation}\label{submodular-formulation}
    \max_{M^D \subseteq R, |M^D| \leq \varrho^D} \mathbb{E}_{\pmb{\lambda} \in p(\pmb{\lambda})}\Bigg[\dfrac{\max_{r \in M^D} s_{\pmb{\lambda}}(u_f(g(r)))}{\max_{r \in R} s_{\pmb{\lambda}} (u_f(g(r)))}\Bigg]
\end{equation}
where we assume $p(\pmb{\lambda})$ is the prior with support $\Lambda$ imposed on the set of DVH metric trade-off values. To solve Equation \ref{submodular-formulation}, we employ a modified version of the dense sampling procedure from \citet{chen_mosh_2024}, where we aim to sample from a discrete and constrained input space instead. We find that doing so prevents exploration of undesirable regions of the input space and greatly improves the computational efficiency, which we highlight further in Section \ref{sec:eval-iterative-exploration}. To further improve computational efficiency, we provide a warm start by first employing a coarse grid search over $R^{\dagger}$, $g(r)$ $\forall r \in R^{\dagger}$, where $R^{\dagger}_k = \{r^{\dagger}_k \mid r^{\dagger}_k = \ell^{\dagger}_k + c\Delta^{\dagger}_k, c \in \mathbb{N}, r^{\dagger}_k \leq u^{\dagger}_k\}$. We then use simple piecewise linear interpolation for a finer-grained input space, $\hat{g}(r)$ $\forall r \in R^{\ddagger}$, where $\ell^{\ddagger}_k = \ell^{\dagger}_k$, $u^{\ddagger}_k = u^{\dagger}_k$, and $\Delta^{\ddagger}_k < \Delta^{\dagger}_k$ $\forall$ $k \in K\backslash \{PTV\}$. The dense set of treatment plans, $M^D$, is then sampled from this input space, $R^{\ddagger}$. The full algorithm is depicted in Algorithm \ref{algorithm:dvh-sampling}.

\begin{algorithm2e}
\caption{MoSH-Dense: Treatment Plan DVH Metric Trade-Offs Sampling Algorithm}
\label{algorithm:dvh-sampling}

Initialize aim and limit values $\{\alpha_{k}^{aim}, \alpha_{k}^{lim}\}$ $\forall$ $k \in [K]$\;
Conduct coarse grid search $g(r)$ $\forall r \in R^{\dagger}$\;
Perform fine-grained piecewise linear interpolation $\hat{g}(r)$ $\forall r \in R^{\ddagger}$\;
Initialize $M^D \leftarrow \emptyset$\;

\While{$|M^D| \leq \varrho^D$}{
    Obtain $\lambda \sim p(\lambda)$\;
    
    $r \leftarrow \operatorname{arg\,max}_{r \in R^{\ddagger}} s_{\pmb{\lambda}}(u_f(\hat{g}(r)))$\;
    
    Obtain $y \leftarrow f(g(r))$ \tcp*[r]{Evaluate treatment plan DVH metrics}
    
    $M^D \leftarrow M^D \cup \{y\}$\;
}
\KwRet{$M^D$}
\end{algorithm2e}

\noindent
\textbf{Step 2: MoSH-Sparse.}
MoSH-Dense provides a dense set of Pareto optimal treatment plans $M^D$. The goal for MoSH-Sparse is now to sparsify $M^D$ such that it provides the clinicians with an easily navigable set of treatment plans, $M^S$, which is still as robust to the worst-case $\pmb{\lambda^*}$. To do so, MoSH-Sparse leverages the concept of submodularity, which here captures the notion of diminishing returns in utility for each additional treatment plan the clinician validates. Equation \ref{eq:submodular-formulation} is then formulated as a robust submodular observation selection (RSOS) problem as such \citep{krause_robust_2008}:
\begin{equation}\label{submodular-formulation2}
    \max_{M^S \subseteq M^D, |M^S| \leq \varrho} \min_{\pmb{\lambda} \in \Lambda}\underbrace{\Bigg[\dfrac{\max_{r \in M^S} s_{\pmb{\lambda}}(u_f(x))}{\max_{r \in M^D} s_{\pmb{\lambda}}(u_f(x))}\Bigg]}_{F_{\pmb{\lambda}}}
\end{equation}
where the term $F_{\pmb{\lambda}}$ is proven to be submodular. To solve Equation \ref{submodular-formulation2}, we directly employ MoSH-Sparse, which is guaranteed to find solutions which are at least as informative as the optimal solution, only at a slightly higher cost: $|M^S| \leq \psi \varrho$, for $\psi=1+\log(\max_{r \in M^D} \sum_{i} F_{i}(\{r\}))$. The full algorithm is presented in Algorithms \ref{algorithm:saturate} and \ref{algorithm:gpc}. The returned set, $M^S$ ($M$), is a sparse set of treatment plans, obtained according to the clinician-defined aim and limit values, which is then displayed to the clinician. Further details may be found in \citet{chen_mosh_2024}.

\begin{algorithm2e}[!t]
\caption{MoSH-Sparse: Treatment Plan Pareto Frontier Sparsification}
\label{algorithm:saturate}

\KwIn{$F_1, \dots, F_{|\Lambda|}, \varrho, \psi$}
$q_{\min} \leftarrow 0$; $q_{\max} \leftarrow \min_i F_i(M^D)$; $M^S_{best} \leftarrow \emptyset$\;

\While{$(q_{\max}-q_{\min}) \geq 1 / |\Lambda|$}{
    $q \leftarrow (q_{\min}+q_{\max})/2$\;
    
    Define $\bar{F}_q(M^S) \leftarrow (1 / |\Lambda|) \sum_i \min \{F_i(M^S), q\}$\;
    
    $\hat{M^S} \leftarrow \text{GPC}(\bar{F}_q, q)$ \tcp*[r]{Algorithm \ref{algorithm:gpc}}
    
    \eIf{$|\hat{M^S}| > \psi k$}{
        $q_{\max} \leftarrow q$\;
    }{
        $q_{\min} \leftarrow q$; $M^S_{best} \leftarrow \hat{M^S}$\;
    }
}
\end{algorithm2e}

\begin{algorithm2e}
\caption{Greedy Submodular Partial Cover (GPC) Algorithm \citep{krause_robust_2008}}
\label{algorithm:gpc}
\KwIn{$\bar{F}_q, q$}
\KwOut{$M^S$, the selected set}

$M^S \leftarrow \emptyset$\;

\While{$\bar{F}_q(M^S) < q$}{
  \ForEach{$m \in M^D \setminus M^S$}{
    $\delta_m \leftarrow \bar{F}_q(M^S \cup \{m\}) - \bar{F}_q(M^S)$\;
  }
  $M^S \leftarrow M^S \cup \{\operatorname{arg\,max}_m \delta_m\}$\;
}
\end{algorithm2e}

This framework is inherently interactive; the aim and limit sliders serve as a feedback mechanism, allowing the system to iteratively infer the clinician's latent preferences by observing how they adjust these soft-hard bounds. We discuss this in Appendix \ref{aim-limit-interaction-section}.

\subsubsection{Multi-Objective Treatment Planning Visualization}\label{visualization-section}

\begin{figure}[h]
    \centering
    \includegraphics[width=0.49\textwidth]{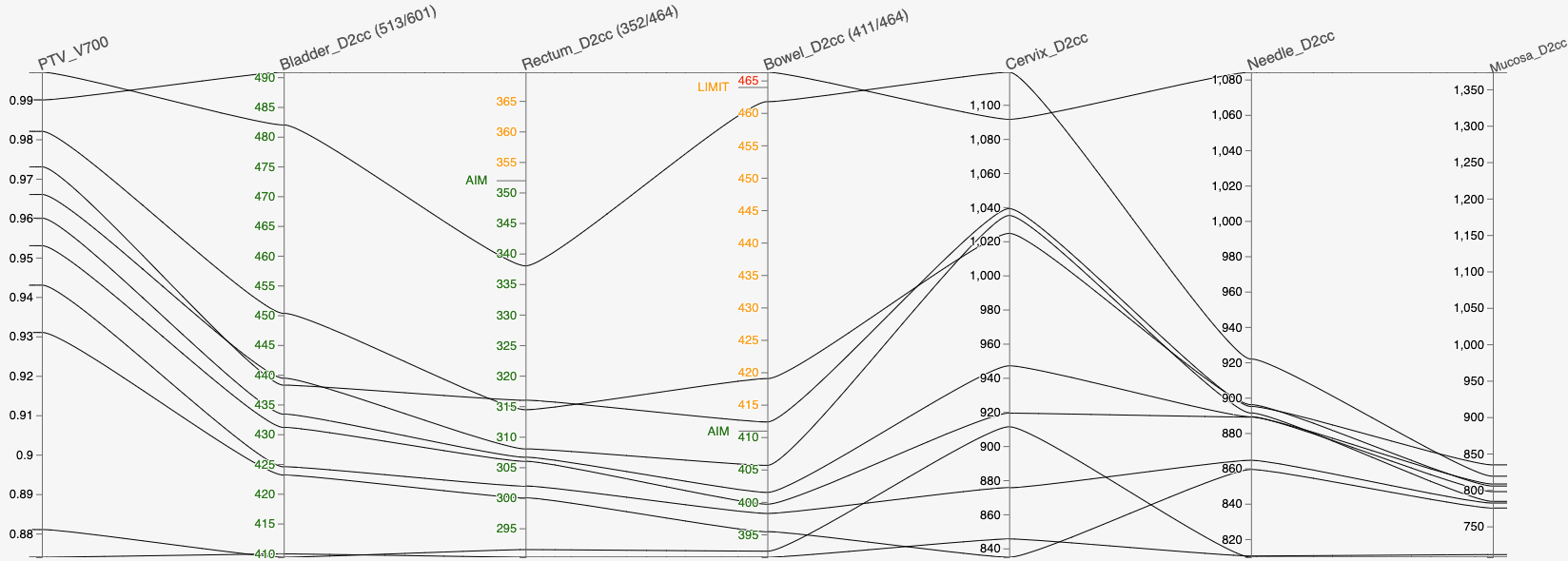}
    \caption{Global Tradeoffs Visualization. Parallel coordinates plot displaying one iteration of treatment plans on the Pareto frontier is shown. Each of the parallel coordinates displays the DVH metrics corresponding to a single treatment plan. This plot is displayed in the center of the tool GUI.}
    \label{fig:parallel-coords}
\end{figure}

\begin{figure}
    \centering
    \includegraphics[width=0.49\textwidth]{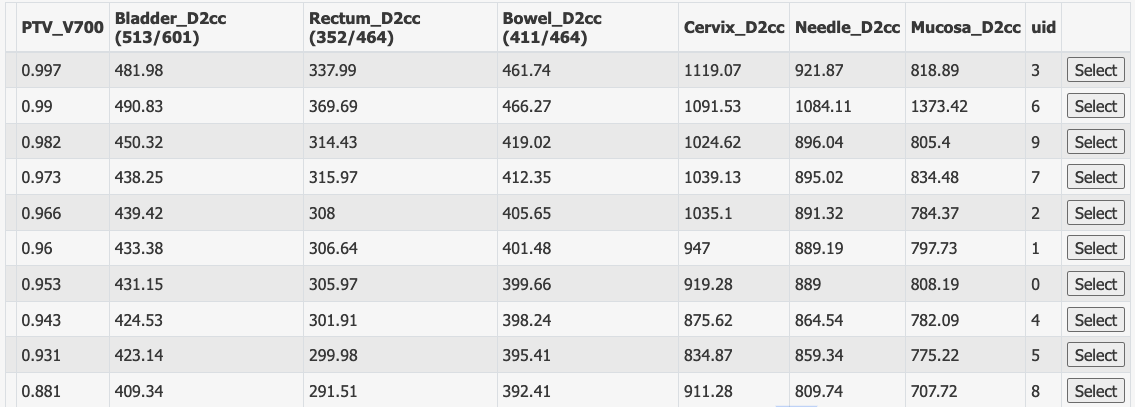}
    \caption{Global Tradeoffs Understanding. Interactive table displaying the dose-volume-histogram (DVH) metrics for the planning target volume (PTV), organs-at-risk (OARs), and all of the regions at risk of hot spots. This table appears directly below Figure \ref{fig:parallel-coords} in the tool GUI.}
    \label{fig:metrics-table}
\end{figure}

\textbf{Global Tradeoffs Understanding.} 
ALMo's visualization framework provides a comprehensive representation of the multi-objective treatment planning space through two complementary views. The system enables intuitive exploration of dosage trade-offs in the DVH metrics space through a parallel coordinates visualization, where treatment plans are displayed, with the clinician-defined aim and limit values clearly delineated (Figure \ref{fig:parallel-coords}). This visualization is supplemented by an interactive tabular interface with the numerical metrics, facilitating precise quantitative assessment (Figure \ref{fig:metrics-table}).

\begin{figure}[ht]
    \centering
    \includegraphics[width=0.495\textwidth]{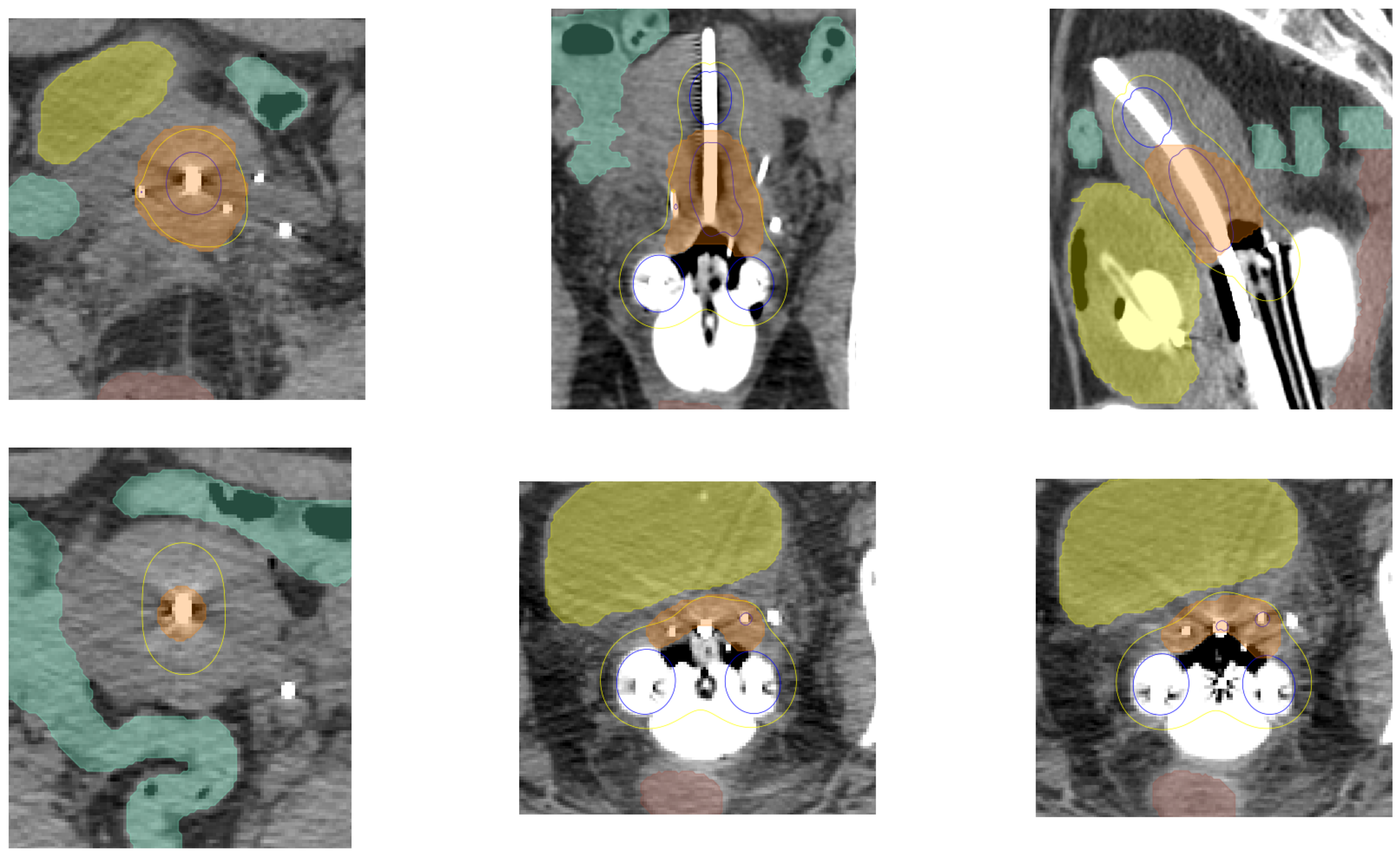}
    \caption{Local Tradeoffs Visualization. Interpretability figures, ALMo-Viz-Explain, for a single treatment plan are shown. The top row displays the most salient views from the axial, sagittal, and coronal perspectives, and the bottom row displays the axial frames with the hottest regions in the cervix, vaginal mucosa, and needle regions. 
    }
    \label{fig:interpretability-figure}
\end{figure}

\noindent
\textbf{Local Tradeoffs Understanding.} 
To facilitate independent operation from conventional treatment planning systems, we developed ALMo-Viz-Explain, an integrated visualization grid that systematically presents critical local dosimetric information. This component streamlines the evaluation workflow by minimizing the need for platform switching during plan assessment. The visualization grid employs a two-tiered structure for comprehensive plan evaluation. The upper tier presents CT slices from anatomically significant viewpoints: axial, sagittal, and coronal planes. Following established clinical protocols for HDR brachytherapy visualization\footnote{Based on standardized treatment planning protocols at the Stanford University Medical Center and aligned with American Brachytherapy Society guidelines.}, we identify key anatomical landmarks: the central tandem applicator plane, the characteristic pear-shaped isodose distribution plane, and the optimal tandem applicator visibility plane -- displayed from the axial, sagittal, and coronal viewpoints, respectively. The lower tier focuses on regions of potential toxicity concern, displaying axial slices corresponding to peak dosage regions in cervical, vaginal mucosa, and needle regions (HSRs). Figure \ref{fig:interpretability-figure} demonstrates the ALMo-Viz-Explain interface, which allows the clinician to take a deeper look into a single treatment plan and quickly glean the critical information.

\subsection{Automated Planning Parameter Setup}\label{preprocessing-section}

The Automated Planning Parameter Setup component establishes the computational foundation of ALMo through three complementary algorithms that optimize the initialization and exploration of the treatment planning space. First, the (1) Reference Plan Initialization Algorithm automates the creation of reference dosage distributions (used in Section \ref{mco-lp-section}), which traditionally requires manual clinician input. Second, the (2) Limiting Organs Prediction Algorithm and (3) Epsilon-Ratio Method work together to constrain the search space of DVH metric trade-offs (Section \ref{mco-lp-section}), significantly reducing computational complexity while maintaining clinical relevance. These algorithms, executed during system initialization, enable efficient exploration of the tradeoff surface by establishing appropriate search space bounds and identifying the critical anatomical structures that most significantly influence the optimization process. 

\subsection{Reduce Manual Input: Reference Plan Initialization Algorithm}\label{ref-plan-init-section}
A dosage reference plan is necessary when utilizing the TCVaR metric, as detailed in Section \ref{app:background-metrics}. Rather than necessitating valuable clinician planning time to manually define a reference treatment plan \citep{deufel_pnav_2020}, we automate this process via the following heuristic: we perform a coordinate-wise search over a range of multipliers exclusively for the HSRs ($R_1 \bigtimes ... \bigtimes R_{|H|}$), imposing the voxel-wise dosage constraint until the DVH metric for a single OAR structure exceeds its respective dosage aim value. We define the voxel-wise dosage constraint as $d_i \leq \tau \cdot r_k$ $\forall$ $i \in V_k \subseteq V$, where $\tau$ is the prescription dosage (700 cGy). The CVaR metric could also be used as it does not require a reference plan.

This termination criterion, stopping immediately upon the first violation of an OAR aim, is strategically designed to identify the "limiting organ," defined here as the specific anatomical structure that most significantly constrains the potential for increases in the PTV dose. By incrementally expanding the high-dose region via the HSRs, this approach effectively probes the anatomical geometry to find the tightest bottleneck. As demonstrated in our results in Section \ref{sec:eval-automated-planning-setup}, any bias towards a specific OAR does not negatively impact the final plan quality, as the subsequent Epsilon-Constraint Method phases are sufficiently robust to refine this initialization and recover any necessary granularity. The resultant dosage distribution from the final iteration is strictly used as the reference dosage distribution, $d^{init}$, for the remainder of the pipeline. The full procedure is detailed in Algorithm \ref{algorithm:reference-plan-init}.

\begin{algorithm2e}
\caption{Automatic Reference Plan Initialization}
\label{algorithm:reference-plan-init}

Initialize $\alpha^{aim}_k$ $\forall$ $k \in K \setminus H$\;
Initialize $R_\eta$ for $\eta \in H$ \tcp*[r]{Cervical, Vaginal Mucosa, and Needle structures}

\For{$r \in R_1 \bigtimes \dots \bigtimes R_{|H|}$}{
    \If{$f_k(g(r)) \geq \alpha^{aim}_k$ for any $k \in K \setminus H$}{
        \KwRet{$g(r)$}
    }
}
\end{algorithm2e}

\subsection{Constrain Search Space: Limiting Organs Prediction Algorithm}\label{limiting-organs-section}
As described in Section \ref{mco-lp-section}, an initial grid search over $R^{\dagger}$ is performed in order to obtain a coarse representation of the high-dimensional Pareto frontier. Due to the combinatorial explosion of the number of parameters in the grid search for the OARs (bladder, rectum, and bowel) along with the HSRs (cervical, vaginal mucosa, and needle regions), we propose to search only over the HSRs and the \textit{limiting} OARs. Without doing so, a grid search over both the OARs and HSRs is computationally prohibitive. We define the set of limiting organ(s) $L$ to be the OARs which more easily satiate their respective [$\alpha^{aim}, \alpha^{lim}$] ranges, due to proximity to the applicators and patient anatomy. We formulate this as follows:

$$L = \argmax_{o \in O}\sum_{r \in R}{\mathbbm{1}\{f_o(g(r)) \geq \alpha_o^{aim}\}}$$

\noindent
where $R = R_1 \bigtimes ... \bigtimes R_{|H|}$ and $O$ is the set of OAR.

We operationalize the above and discover the limiting organs by performing a coordinate-wise search over the HSRs until one (or more) of the DVH metrics of the OAR structures surpasses its respective aim value, similar to what is described in Algorithm \ref{algorithm:reference-plan-init}. In cases where multiple of the OARs surpass their respective aim values in the same iteration, there would be more than one limiting organ used. \textit{A differentiating factor between this algorithm and Algorithm \ref{algorithm:reference-plan-init} is that we do not use the voxel-wise dosage constraint and instead use the TCVaR approximations in Equation \ref{equation:moo}. This is done to obtain a better approximation of the limiting OARs.}

\subsection{Constrain Search Space: Epsilon-Ratio Method}\label{epsilon-ratio-section}
As it is computationally infeasible to efficiently compute a grid search covering the entire space of possible DVH metric trade-offs, ALMo focuses only on the region of the Pareto frontier which encompasses the reasonable range of DVH values for each of the OARs (in this case, the respective aim and limit values for each OAR). As described in Section \ref{mc-modeling-section}, we aim to determine the lower and upper bounds, $\ell^{\dagger}_k$ and $u^{\dagger}_k$ $\forall$ $k \in K$, which correspond to a reasonable range of DVH values when defining the input space $R^{\dagger}$. The challenge with this arises from the fact that we are not aware of the $r_k$ multipliers, when solving Equation \ref{equation:moo}, necessary to attain the reasonable range of DVH values for each OAR. As a result, we want to derive at least an approximate relationship between the DVH metric and its convex approximation, the TCVaR, for each region. In particular, we seek to determine the relationship between the parameter $\varepsilon_k$, which constrains the TCVaR of the $k$th region of interest, and the aim and limit, $\alpha^{\text{aim}}_k$ and $\alpha^{\text{lim}}_k$, respectively, for the DVH metric of the $k$th region of interest. 

Ideally, we want to find some transformation $T$ such that setting $\hat{r}_k = T(\alpha_k)$ and constraining the convex optimization to $\theta^{+}_k(\cdot) \leq \hat{r}_k * \gamma_k^{+, init}$ closely approximates $f_k(g(\hat{r})) \simeq \alpha_k$. We derive this transformation, using what we refer to as the Epsilon-Ratio Method, by performing 
\begin{equation}
    \hat{r}^{aim}_k = \dfrac{\alpha^{\{aim\}}_k}{f_k(d^{init})} \cdot \gamma^{+,init}_k
\label{equation:epsilon-ratio}
\end{equation}
The Epsilon-Ratio Method provides us with a close approximation which we iteratively adjust until we find a satisfactory $\hat{r}_k$ for each $k \in K$ for the grid search. Prior to computing $R^{\dagger}$ (Section \ref{mc-modeling-section}), we then set $\ell_k = \hat{r}^{aim}_k$, $u_k = \hat{r}^{lim}_k$ $\forall$ $k \in K\backslash\{PTV\}$. The algorithm details are shown in Algorithm \ref{algorithm:epsilon-ratio-method}.

\begin{algorithm2e}
\caption{Epsilon-Ratio Method}
\label{algorithm:epsilon-ratio-method}

Obtain $f_k(d^{init})$, $\gamma^{+,init}_k$ $\forall k \in K \setminus H$ from reference distribution $d^{init}$\;
Calculate $\hat{r}_k$ $\forall k \in K \setminus H$ for both aim and limit using Equation \ref{equation:epsilon-ratio}\;
Initialize step size $c$\;

\For{$\alpha_k \in \{\alpha^{aim}_k, \alpha^{lim}_k\}$ $\forall k \in K \setminus H$}{
    \While{$f_k(g(\hat{r}_k)) \notin [\alpha_k-\epsilon, \alpha_k+\epsilon]$}{
        \eIf{$f_k(g(\hat{r}_k)) \leq \alpha_k$}{
            $\hat{r} \leftarrow \hat{r} + c$\;
        }{
            $\hat{r} \leftarrow \hat{r} - c$\;
        }
    }
}
\end{algorithm2e}

\section{Results}\label{sec3}
All experimental setup details are provided in Appendix \ref{sec:experiental-setup}. 

\subsection{Treatment Plan Optimization}\label{sec:experiment-treatment-plan}

To evaluate the quality of the treatment plans from Section \ref{inner-loop-section}, we aimed to show that our proposed Equation \ref{equation:moo} is (a) capable of producing clinically viable treatment plans, and (b) capable of matching or exceeding the DVH metrics of the treatment plans manually planned in the clinic. We first retrospectively obtained 20 patient cases from the clinic and worked with clinicians to devise a grading scheme which assigns numerical scores to treatment plans. We aimed to take into consideration the following factors: clinical viability, hot spots toxicity levels, dosage distribution shapes, and relative improvements in DVH metrics compared to those of the clinically treated plan. The grading scheme is shown in Figure \ref{fig:plan-grading-scheme}.

\begin{figure}[ht]
    \centering
    \begin{tcolorbox}
    [colback=blue!5!white,title=Treatment Plan Grading Scheme (Relative To Clinical Treatment Plan)]\label{plan-grading-scheme}
        Grade 3: Major Deficiencies
        \begin{itemize}
            \item 200\% of prescription dosage within the vaginal mucosa region
            \item Only PTV is treated, no pear-shaped dosage distribution
            \item Limits (not aim) exceeded on OARS
        \end{itemize}
        Grade 2: Minor Deficiencies
        \begin{itemize}
            \item 200\% of prescription dosage outside of the PTV
            \item No improvement in PTV coverage and OAR metrics
        \end{itemize}
        Grade 1: Clinically Acceptable with Minor Improvements
        \begin{itemize}
            \item Minor improvements in PTV coverage [0.5-3\%] or OAR metrics [20-50 cGY], while keeping the other, PTV coverage or OAR metrics, within 0.5\% and 20 cGY, respectively
        \end{itemize}
        Grade 0: Clinically Acceptable with Major Improvements
        \begin{itemize}
            \item Major improvements in PTV coverage [$>3\%$] or OAR metrics [$>50$ cGY], while keeping the other, PTV coverage or OAR metrics, within 0.5\% and 20 cGY, respectively
        \end{itemize}
    \end{tcolorbox}
    \caption{Treatment Plan Grading Scheme. Treatment plan grading scheme designed through discussions with clinicians and used for evaluating treatment plans shown in Tables \ref{table:plan-grades} and \ref{table:iterative-plan-grades}.}
    \label{fig:plan-grading-scheme}
\end{figure}

For each clinical case, we used ALMo to generate three distinct treatment plans with varying DVH metrics. We then asked a clinician to identify and grade the plan that best aligned with their internal preferences. Planning time was excluded from this analysis to focus solely on evaluating the capabilities of our proposed method. As a baseline, we used the optimization method from \citet{deufel_pnav_2020}. While this approach does not explicitly control for hot spot toxicity, we included it to empirically demonstrate the necessity of such constraints for generating clinically viable plans. Table \ref{table:plan-grades} illustrates our results.

\begin{table}[t]
    \centering
    \begin{tabular*}{\columnwidth}{p{0.55\columnwidth} @{\extracolsep{\fill}} c c c c} 
      \toprule
      \textbf{Method} & \textbf{G0} & \textbf{G1} & \textbf{G2} & \textbf{G3} \\ 
      \midrule
      Ours & 3 & 10 & 7 & 0\\
      \midrule
      PNaV Treatment Plan \citep{deufel_pnav_2020} & 0 & 0 & 2 & 18\\
      \bottomrule
    \end{tabular*}
    \caption{Treatment Plan Optimization Results. Treatment plan grading results for 20 cases based on our clinician-devised grading scheme (Figure \ref{fig:plan-grading-scheme}). Grades refer to Grade 0 (G0) through Grade 3 (G3). For each case, three different plans were selected and the clinician graded the plan closest to their internal preferences. This evaluates the optimization component independent of planning time.}
    \label{table:plan-grades}
\end{table}

As shown in Table \ref{table:plan-grades}, ALMo is capable of producing treatment plans with minimal to no clinical deficiencies, often with DVH metrics which greatly improve upon those of clinical treatment plans. The results further illustrate the importance of explicitly controlling for hot spots toxicity levels, which the baseline method fails to account for -- highlighting the clinical practicality of our method.

\subsection{Iterative Clinician Exploration}\label{sec:eval-iterative-exploration}

\subsubsection{End-to-End Evaluation}\label{sec:e2e-iterative-eval}
We retrospectively obtained another set of five patient cases from the clinic (different from those in Section \ref{sec:experiment-treatment-plan}) and conducted the following experiment. For each patient case, we conducted three treatment planning trials utilizing the Iterative Clinician Exploration component. In each trial, we iteratively adjusted the aim and limit values based on observed treatment plans at each iteration and stopped once we were satisfied with the displayed plans. Specifically, for each patient case, we: (1) selected three treatment plans from distinct regions of the Pareto frontier, (2) evaluated the one plan most similar to the clinical treatment plan based on DVH metrics, using our standardized grading scheme (Figure \ref{fig:plan-grading-scheme}), and (3) documented the mean planning duration. The results are presented in Table \ref{table:iterative-plan-grades}. As shown, the Iterative Clinician Exploration component allows for ALMo to consistently produce treatment plans with minimal to no clinical deficiencies, often with minor DVH metric improvements over the clinical treatment plans. The average planning time of 17 minutes demonstrates an improvement over traditional approaches ($\sim$ 30-90 minutes) \citep{michaud_workflow_2016, hansen_comparing_2010, oud_fast_2020}. Appendix \ref{sec:iterative-viz-eval} describes results for evaluation of the ALMo-Viz-Explain visualization component. Experimental setup details are described in Appendix \ref{sec:experiental-setup}.

\begin{table}[t]
    \centering
    \begin{tabular*}{\columnwidth}{l @{\extracolsep{\fill}} c c c c p{0.35\columnwidth}} 
      \toprule
      \textbf{Method} & \textbf{G0} & \textbf{G1} & \textbf{G2} & \textbf{G3} & \textbf{Planning Time} \\ 
      \midrule
      ALMo & 0 & 3 & 2 & 0 & 17.6min $\pm$ 1min 43sec\\
      \bottomrule
    \end{tabular*}
    \caption{Iterative Clinician Exploration Planning Results. Results for five cases based on our clinician-devised grading scheme (Figure \ref{fig:plan-grading-scheme}), where G0 is the best grade and G3 is the worst. We selected three treatment plans from distinct regions of the Pareto frontier and graded the plan most similar to the clinical treatment plan. The mean planning time was averaged across all five patient cases.}
    \label{table:iterative-plan-grades}
\end{table}

\subsubsection{Treatment Plan Exploration Efficiency }\label{sec:treatment-plan-exploration-efficiency}
We next evaluated the Iterative Clinician Exploration component to assess the diversity of its generated treatment plans. For each patient case, we compared our proposed sampling method (Algorithm \ref{algorithm:dvh-sampling}) against an exhaustive grid search over the refined parameter space, $R^{\ddagger}$ (see Section \ref{mc-modeling-section}). The objective was to verify that our component efficiently delivers a broad set of Pareto-optimal plans. Ideally, these plans should provide high coverage of the dosage tradeoff surface, strictly bounded by the clinician's defined aim and limit values.

\begin{figure}[ht]
\begin{center}
\includegraphics[width=0.499\textwidth]{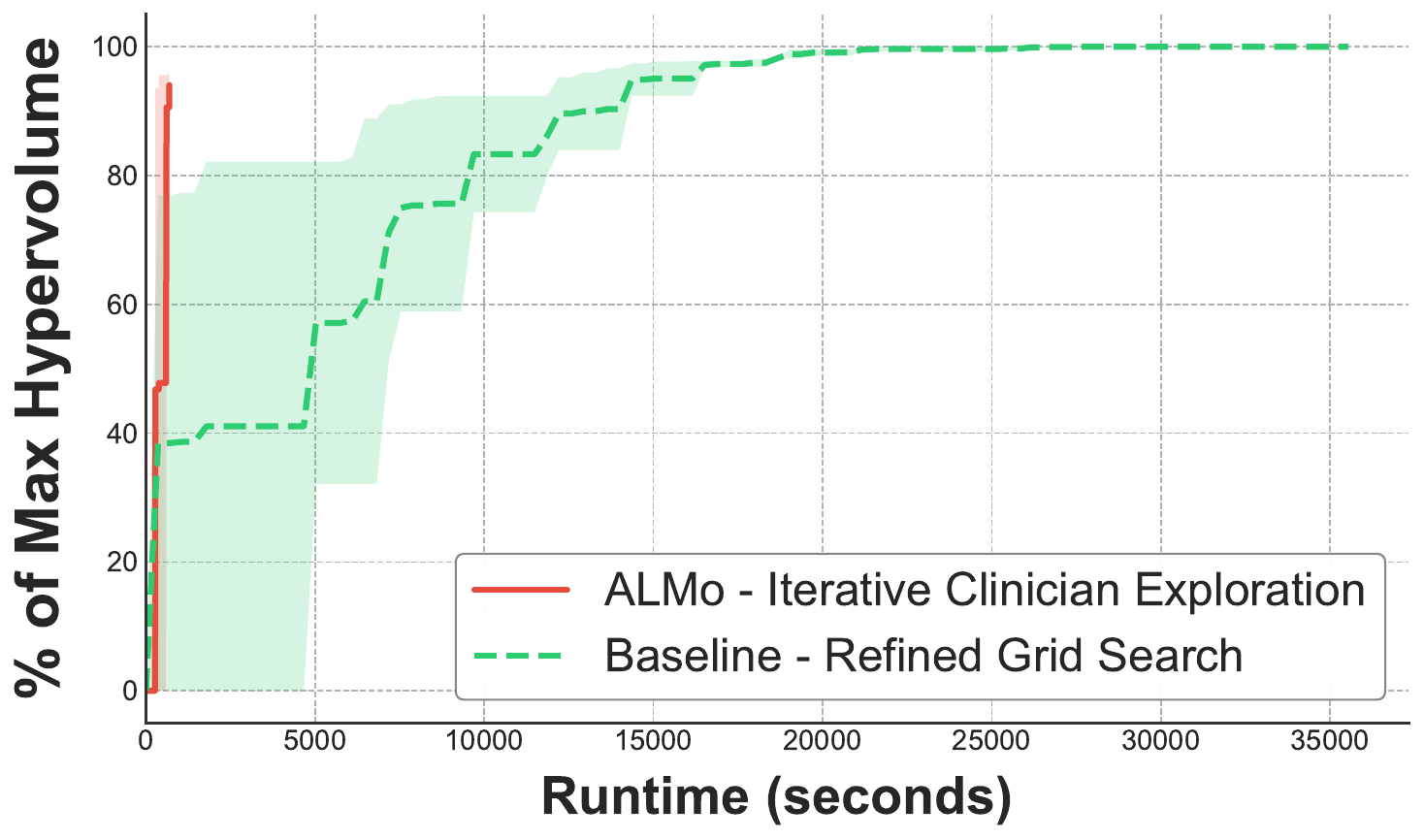}
\end{center}
\caption{Iterative Clinician Exploration Hypervolume Results. Plot comparing the Iterative Clinician Exploration component of ALMo against a refined grid search baseline via the hypervolume metric for each timestamp (in seconds). The hypervolume is calculated with respect to the set of limit values described in Appendix \ref{sec:experiental-setup} and is represented as a percentage value of the maximum hypervolume metric for each patient case. The results are averaged across five patient cases and the darker solid line depicts the mean values, whereas the lighter shading depicts the standard deviation values.}
\label{fig:hypervolume-by-runtime}
\end{figure}

To quantitatively measure what we want, we employed the hypervolume indicator metric (Section \ref{sec:moo-background}) with respect to the clinician-defined limit values at each timestamp (i.e. the hypervolume indicator reference point is set to the limit values described in Appendix \ref{sec:experiental-setup}). The results, averaged across the five patient cases, are shown in Figure \ref{fig:hypervolume-by-runtime}. As shown, the Iterative Clinician Exploration component proves to reach $\sim$ 95\% of the max hypervolume, on average, 14x faster (in computational wall-clock time) than a complete grid search over the refined set of parameters $R^{\ddagger}$. This acceleration is driven by the system's ability to efficiently sample the relevant solution space without the exhaustive evaluation required by grid search, further highlighting the efficiency improvements provided by ALMo.

\subsubsection{ALMo-Viz-Explain Evaluation}\label{sec:iterative-viz-eval}
The Iterative Clinician Exploration module incorporates ALMo-Viz-Explain, a grid-based visualization framework designed to highlight critical dosimetric features and facilitate granular local dosage analysis. Since ALMo functions independently of standard commercial platforms, this integrated tool aims to streamline the evaluation workflow by reducing the cognitive load and time overhead associated with toggling between disparate software systems.

To quantify the utility of this integration, we monitored the frequency with which treatment plans could be assessed exclusively via ALMo-Viz-Explain, without recourse to conventional dosage visualization software (e.g., the ARIA Oncology Information System; see Section \ref{sec:experiental-setup}). Across the five clinical cases described in Section \ref{sec:eval-iterative-exploration}, involving the evaluation of 15 distinct treatment plans, external software was required for only 5 plans.

Qualitatively, we observed that ALMo-Viz-Explain was particularly effective for the rapid rejection of non-viable plans, as it facilitated the immediate identification of unfavorable dose distribution shapes and excessive hot spots. Conversely, plans that appeared clinically viable typically prompted a secondary, comprehensive review using conventional software to ensure strict safety compliance. Ultimately, the integrated visualization proved sufficient for decision-making in 66.7\% of the evaluated iterations (10 out of 15 plans), a factor that significantly contributed to the planning efficiency improvements reported in Section \ref{sec:e2e-iterative-eval}.

\subsection{Automated Planning Parameter Setup}\label{sec:eval-automated-planning-setup}

We validate our three proposed algorithms in this section, where the Reference Plan Initialization Algorithm aims to compute the central treatment plan for use in calculating the TCVaR metric (Equation \ref{eq:tcvar-metric}), and the Limiting Organs Prediction Algorithm and Epsilon-Ratio Method aim to reduce the dimensionality of and constrain the input space for the grid search used in our proposed Algorithm \ref{algorithm:dvh-sampling}.

\subsubsection{Reference Plan Initialization Algorithm}\label{sec:eval-ref-plan-init}
As described in Section \ref{ref-plan-init-section}, the purpose of the Reference Plan Initialization Algorithm is to compute a central treatment plan upon which subsequent optimization steps are built. This is achieved by solving Equation \ref{equation:moo} using neighboring parameters. While prior works have relied on a manually created treatment plan for this step, we aimed to automate the process via our proposed Algorithm \ref{algorithm:reference-plan-init}. To validate this approach, we retrospectively compared the DVH metric results of our algorithm against those of five real cervical cancer brachytherapy clinical cases using the utility ratio, defined as $f_k(d^{init})/f_k(d^{clinical})$ for the PTV ($k=1$), and $f_k(d^{clinical})/f_k(d^{init})$ otherwise. Consequently, a utility ratio value $> 1$ is desired.

\begin{figure}[h]
\begin{center}
\includegraphics[width=0.49\textwidth]{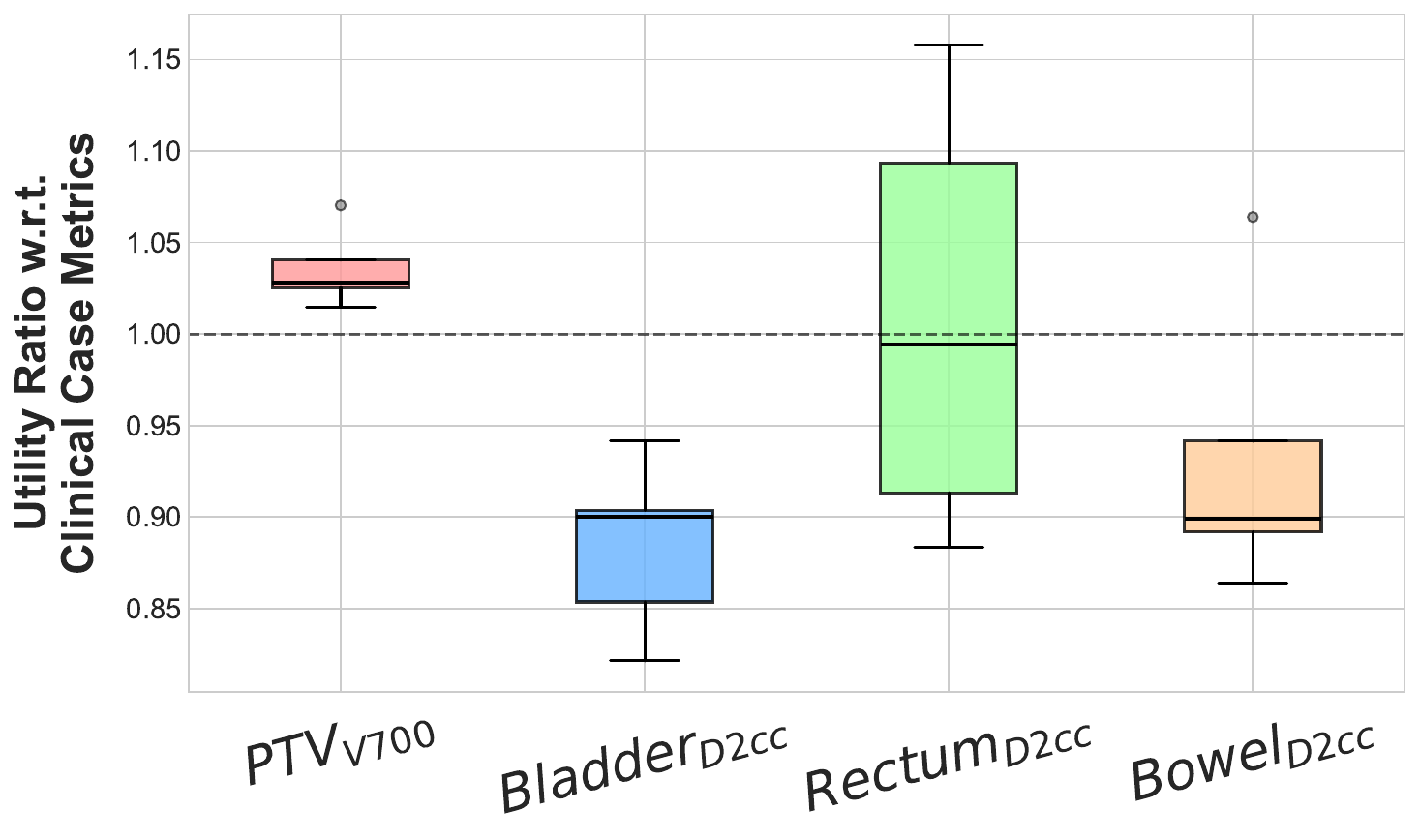}
\end{center}
\caption{Reference Plan Initialization Algorithm Results. Box plot illustrating the utility ratio, based on the DVH metrics, obtained by our proposed Reference Plan Initialization Algorithm, with respect to five manually planned clinical plans. For each DVH metric, a utility ratio value greater than 1 indicates that our Reference Plan Initialization Algorithm obtains treatment plans with improved values (higher for $\text{PTV}_{V700}$ and lower for $\text{Bladder}_{D2cc}$, $\text{Rectum}_{D2cc}$, and $\text{Bowel}_{D2cc}$).}
\label{ref_plan_init_plot}
\end{figure}

Figure \ref{ref_plan_init_plot} illustrates our results. Overall, we observe that Algorithm \ref{algorithm:reference-plan-init} often provides greater utility for the PTV, but lower utility for OARs such as the Bladder and Rectum. These lower utility values may be attributed to the granularity of the multipliers selected for the initialization search space (see Appendix \ref{sec:experiental-setup}). To adhere to the strict time-sensitivity constraints of intraoperative treatment planning, we utilized a coarser granularity for the multiplier search. In practice, while a finer-grained search would likely yield higher initial utility values for structures like the Bladder and Rectum, it would incur a substantial cost in setup time. We view this resolution as a configurable trade-off to be determined by clinical priority.

Crucially, we observed that these lower initial utility values did not negatively impact the final performance of the system. The subsequent application of the $\epsilon$-ratio method effectively compensates for initialization discrepancies by conducting a wider search over the constraint space. For instance, despite the initial lower utility for the Bladder, the Epsilon-Ratio Method successfully determined the appropriate search range, recovering optimal dosimetric values in the final output. This implies that a "warm start" need only be approximately correct, provided the secondary optimization stage—which possesses its own tunable granularity trade-offs—retains sufficient scope to refine the solution. Thus, while Algorithm \ref{algorithm:reference-plan-init} produces reasonable baseline plans, the integration of these additional optimization steps is essential for generating clinically viable plans that support robust decision-making. We validate these additional steps next.

\subsubsection{Limiting Organs Prediction Algorithm}\label{sec:eval-limiting-organs}
\begin{figure}[h]
\begin{center}
\includegraphics[width=0.499\textwidth]{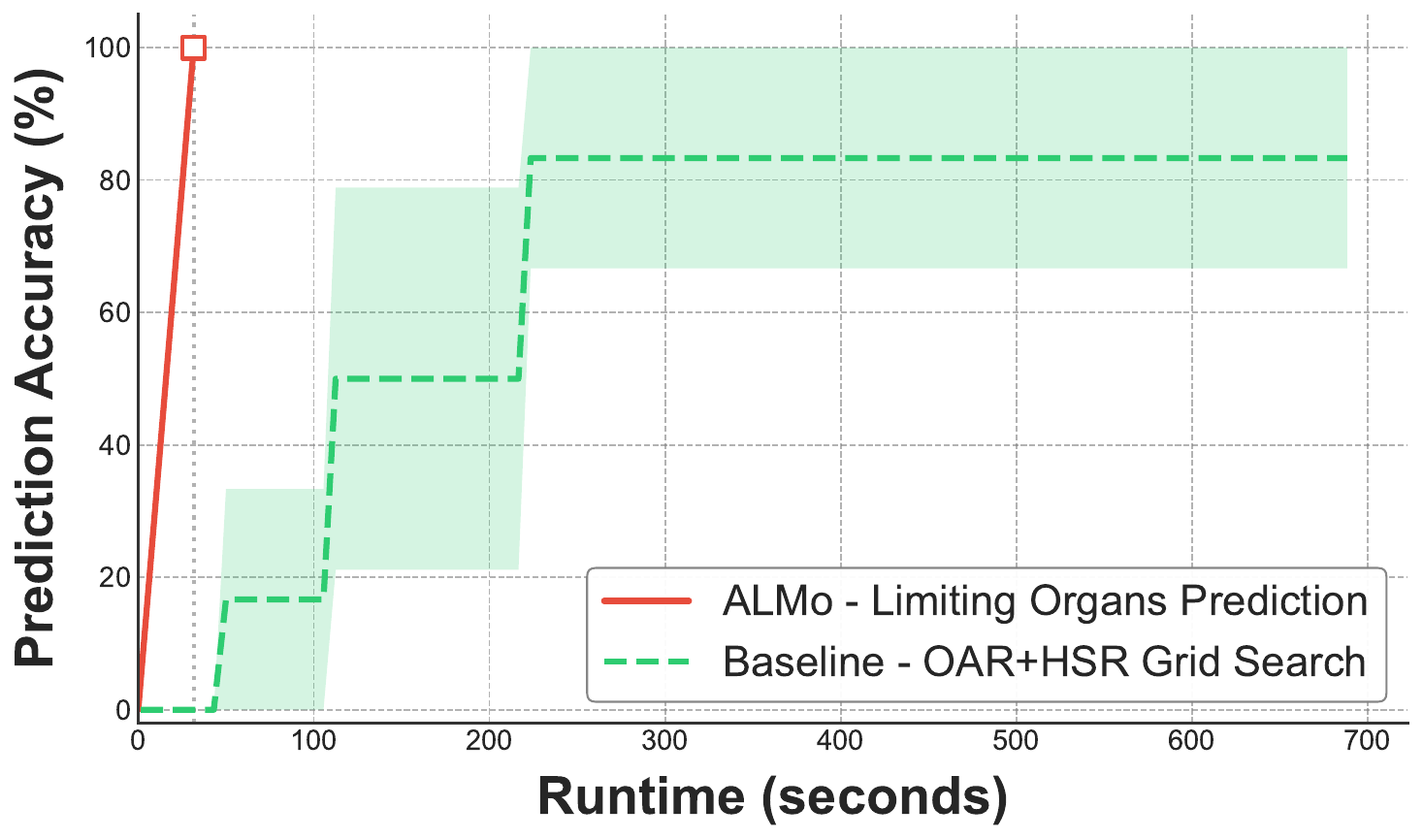}
\end{center}
\caption{Limiting Organs Prediction Algorithm Results. Plot illustrating the results of our runtime vs. prediction accuracy comparison for our proposed Limiting Organs Prediction Algorithm. The results, all averaged over five cases, are compared against an iteratively refined grid search algorithm, shown in green, which was performed using varying resolution values of the grid search space. The darker solid line depicts the mean value, whereas the lighter shading depicts the standard error values. For our proposed Limiting Organs Prediction Algorithm, only the mean runtime and accuracy is illustrated since it obtained 100\% accuracy for each of the five cases.}
\label{limiting_organs_results_plot}
\end{figure}

As described in Section \ref{limiting-organs-section}, searching only over the HSRs and the \textit{limiting} OARs helps to reduce the dimensionality of the input space, $R^{\dagger}$, and make it more computationally feasible. To validate our proposed algorithm, we compared the predicted set of limiting organs from our algorithm, which only performs a coordinate-wise search over the HSRs, against that of a baseline grid search algorithm which performs a search over both the OARs and HSRs. The baseline grid search algorithm was computed over $\bigtimes_{k=2}^{|K|} R^{base}_k$, where $R^{base}_k = \{r^{base}_k \mid r^{base}_k = \ell^{base}_k + c\Delta^{base}_k, c \in \mathbb{N}, r^{base}_k \leq u^{base}_k\}$. We varied the resolution $\Delta^{base}_{2,..,K}$ over \{0.2, 0.22, ..., 0.8\}, kept equal for $k=2,...,K$. As a result, $\Delta^{base} = 0.2$ corresponds to the longest runtime and $\Delta^{base} = 0.8$ corresponds to the shortest runtime. We evaluated this on five clinical cases and found that our proposed algorithm obtained 100\% accuracy in 31.6 seconds, on average. This was about $\sim$ 85\% faster than the earliest, out of the five cases, that the baseline grid search algorithm was able to obtain 100\% accuracy, illustrating the efficiency improvements provided by our Limiting Organs Prediction Algorithm. The results are illustrated in Figure \ref{limiting_organs_results_plot}.

\subsubsection{Epsilon-Ratio Method}\label{sec:eval-epsilon-ratio}
\begin{figure}[ht]
\begin{center}
\includegraphics[width=0.499\textwidth]{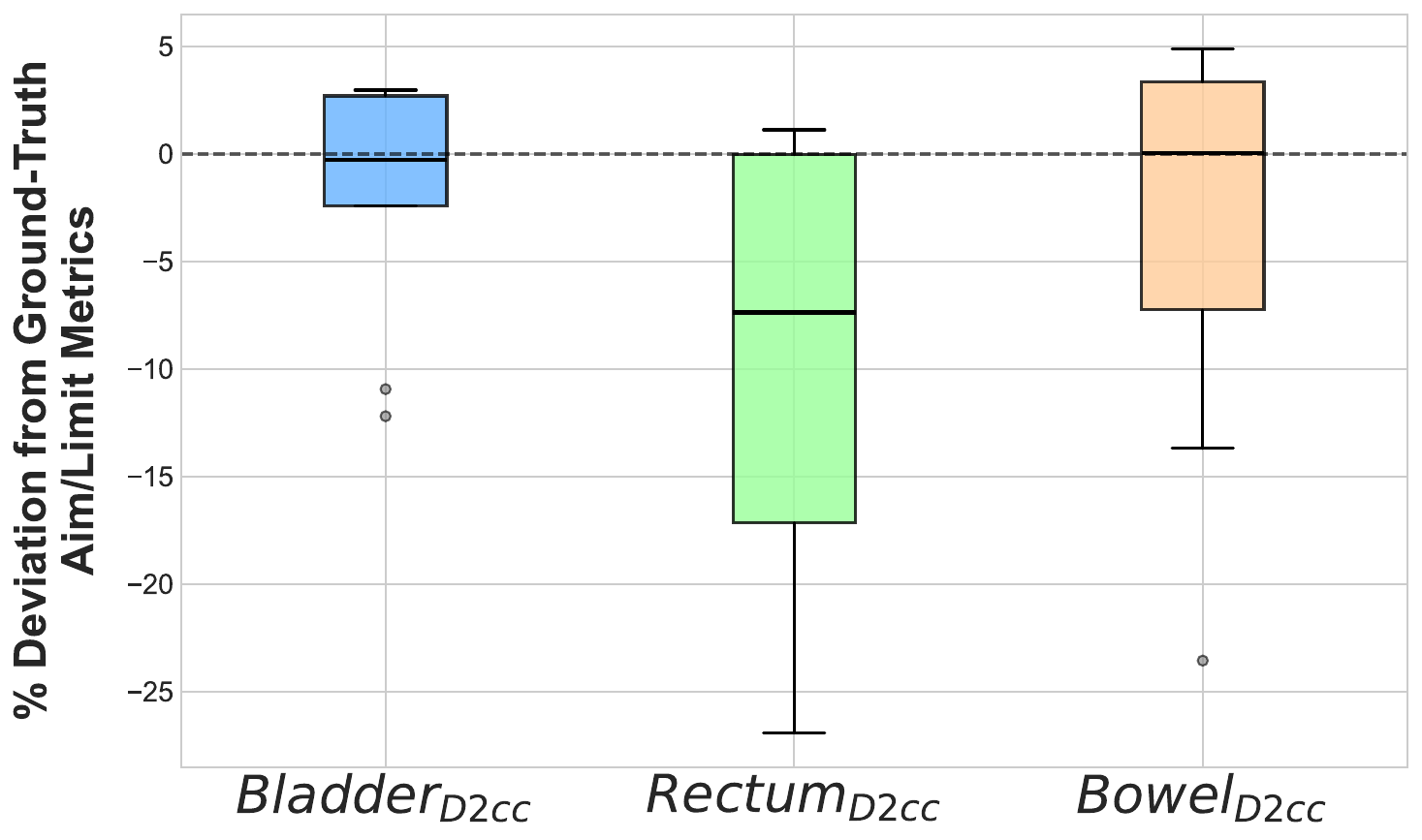}
\end{center}
\caption{Epsilon-Ratio Method Results. Box plot illustrating the percentage deviation of the DVH metrics which result from our estimated $\hat{r}$ parameters, using the Epsilon-Ratio Method, compared to their respective true aim and limit DVH metrics. This is shown for both the aim and limit values across five patient cases.}
\label{epsilon_ratio_plot}
\end{figure}

As described in Section \ref{epsilon-ratio-section}, the Epsilon-Ratio Method focuses the search on regions likely to yield reasonable DVH metrics, effectively bounding the space around the standard aim and limit values defined in Appendix \ref{sec:experiental-setup}. To validate this, we measured the deviation of our estimated metrics (derived from Equation \ref{equation:moo} with $\hat{r}_k$) from the ground-truth values using the formula $(f_k(g(\hat{r})) - \alpha_k^{aim,lim})/\alpha_k^{aim,lim}$. Results across five cases, shown in Figure \ref{epsilon_ratio_plot}, indicate that our estimates typically remain within 8\% of the target values. Overall, they often exhibit a consistent negative skew, indicating that our estimated metrics are often lower than the ground-truth thresholds. This behavior is clinically advantageous: since the optimization is primarily governed by a single limiting OAR (most frequently the Bladder in these samples), which our method characterized accurately, underestimations in non-limiting OAR structures do not compromise the final solution. Consequently, we found the Epsilon-Ratio Method to be highly effective at constraining the search space to relevant regions, thereby improving end-to-end computational efficiency without sacrificing plan quality. 

\section{Discussion, Limitations, and Conclusion}\label{sec4}

Our comprehensive evaluation demonstrates that ALMo successfully addresses critical challenges in HDR brachytherapy treatment planning by integrating a clinically intuitive aim-limit framework with robust plan optimization and setup. The system delivered improvements across three key domains: \textbf{Quality}, \textbf{Efficiency}, and \textbf{Interaction}.

Plan \textbf{quality} was enhanced through a novel approach to hot spots toxicity control. This method resulted in clinically viable plans with no major deficiencies in 100\% of evaluated cases, and 65\% of plans selected with ALMo showed minor to significant dosimetric improvements over their manually created counterparts. Regarding \textbf{efficiency}, ALMo reduced the average end-to-end planning time to approximately 17 minutes, an improvement over conventional approaches. This was driven by algorithmic modifications, including an optimized initialization phase that minimized manual setup and rapid per-plan optimization. Finally, clinician \textbf{interaction} was improved by providing an intuitive interface for navigating the dosage tradeoff surface through its aim-limit guided framework. Our integrated visualization tools also proved sufficient for plan assessment in 66.7\% of cases, significantly reducing the need for external software.

Despite these promising results, certain limitations warrant consideration. First, ALMo is designed as a specialized planning optimization tool rather than a comprehensive end-to-end platform; it does not currently encompass upstream tasks, such as automated image segmentation, or downstream physical quality assurance processes. Second, while our retrospective analysis indicates significant efficiency gains, these metrics await validation in a live intraoperative setting. Real-world clinical deployment introduces dynamic complexities—ranging from user proficiency curves to institutional workflow variability—that must be navigated to confirm the realizable time savings.

Ultimately, these results validate ALMo’s potential to advance personalized brachytherapy by improving clinical workflows and treatment plan standards. Future work will focus on integrating this system into live clinical decision-making environments and extending the framework to other treatment sites and modalities.

\bibliography{chil-sample,references}

\appendix

\section{Appendix}\label{apd:first}

\subsection{Metrics}\label{app:background-metrics}
\noindent
Let $f = (f_1,...,f_{|K|}): \mathbb{R}_+^{|V|} \rightarrow \mathbb{R}_+^{|K|}$ be a vector-valued function of $d \in D$, with $|K|$ ($|K| \geq 2$) components. Each of the $|K|$ components, $f_k$, represents a dosimetric metric of interest and $K$ is the set of structures we optimize over. The dosimetric metrics of interest are known as dose-volume-histogram (DVH) metrics, or value-at-risk (VaR) metrics, e.g. $\text{PTV}_{V700}$, the percentage of radiation dosage coverage, at the prescription dosage (700 cGY), which the PTV receives, or $\text{Bladder}_{D2cc}$, the dosage to the hottest dose to 2 $\text{cm}^3$ of tissue for the bladder. However, due to the non-convexity of VaR metrics, we use convex approximations of the VaR metrics, known as truncated conditional value-at-risk (TCVaR) metrics, for the optimization of each treatment plan \citep{wu_new_2020}. The TCVaR metric was first proposed by \citet{wu_new_2020} as an improvement to the conditional value-at-risk (CVaR) metric. Whereas the VaR metrics measure the radiation dosage value at a given quantile $\beta_k \in [0,1]$ in the distribution, the CVaR metric measures the average dose of the tail of the distribution, specified by some given quantile $\beta_k$. The TCVaR metric, on the other hand, excludes the hottest 100 $\tau_k\%$ from the CVaR calculation, to better approximate the VaR value. We denote the $(\beta_k, \tau_k)$-TCVaR metric for the high-dose tail of the dosage distribution with $\theta_k^+(d;\Bar{d},\beta_k,\tau_k)$, where $\Bar{d} \in \mathbb{R}_+^{|V|}$ is a reference dosage distribution.  

Let $V_k \subseteq V$ represent the set of voxels relevant for structure $k$. We will only describe the aforementioned metrics for the high-dose tail of the dosage distribution below. For $\beta_k \in [0,1]$, $\beta_k$-VaR for voxels $V_k$ is defined as:
\begin{equation}
\label{equation:var}
  \delta_k^+(d;\beta_k) = \min_{\nu_k} \{\nu_k \in \mathbb{R}: \dfrac{|j \in V_k: d_j \geq \nu_k|}{|V_k|} \leq \beta_k\}  
\end{equation}
Intuitively, $\delta_k^+(d;\beta_k)$ is the dosage value to the $|V_k|\beta_k$th hottest voxel. From now on, we will denote the $\beta_k$ used in $\beta_k$-VaR as $\Bar{\beta}_k$. The $\beta_k$-CVaR is defined as:
\begin{equation}
    \gamma_k^+(d;\beta_k) = \min_{\nu_k \in \mathbb{R}}\{\nu_k + \dfrac{1}{\beta_k |V_k|} \sum_{j \in V_k} max(0, d_j-\nu_k) \}
\end{equation}
Intuitively, the $\beta_k$-CVaR represents the average radiation dosage to the hottest 100$\beta_k\%$ of the voxels $V_k$, and is jointly convex in $d$ and $\nu_k$. To define the TCVaR metric, let $\Bar{d} \in \mathbb{R}_+^{|V|}$ be a reference dosage distribution and let $\tau_k \in [0,\beta_k]$. The $(\beta_k, \tau_k)$-TCVaR metric is defined as:
\begin{equation}\label{eq:tcvar-metric}
\begin{split}
    \theta_k^+(d;\Bar{d},\beta_k,\tau_k) = \min_{\nu_k \in \mathbb{R}} \bigg\{ \nu_k + \dfrac{1}{(\beta_k-\tau_k)|V_k|} \\
    \cdot \sum_{\substack{j \in V_k : \\ \Bar{d_j} \leq \delta^+(\Bar{d};\tau_k)}} \max(0, d_j-\nu_k) \bigg\}
\end{split}
\end{equation}
Intuitively, the $(\beta_k, \tau_k)$-TCVaR metric, which we refer to as $\theta_k$, tries to exclude the hottest 100$\tau_k \%$ of the voxels from the radiation dosage distribution and then calculates the CVaR metric using the remaining values. If $\tau_k=0$, then $(\beta_k, \tau_k)$-TCVaR = $\beta_k$-CVaR. If $\Bar{d}=d$ (or if we know exactly which voxels form the $\beta_k$ tail in each structure), and we use $\tau_k = \beta_k = \Bar{\beta}_k$, then $(\beta_k, \tau_k)$-TCVaR = $\Bar{\beta}_k$-VaR. However, we do not know $d$ \textit{a priori}. To resolve that, we follow \cite{deufel_pnav_2020} and use a contracting window iterative algorithm to gradually increase $\tau_k$ and gradually identify which voxels should be truncated to best approximate $d$. In short, the contracting window iterative algorithm starts with $\tau_k = 0$ before iteratively optimizing then increasing it to approximate $d$ (more details may be found in \citet{deufel_pnav_2020}). Unlike \citet{deufel_pnav_2020}, which assumes clinician manual setup of the reference plan $\Bar{d}$, we propose an algorithm to automate the creation of the reference treatment plan, in Section \ref{ref-plan-init-section}. We use the $(\beta_k, \tau_k)$-TCVaR metric for the $k$ structures we care about due to its convexity and potentially better approximation of the non-convex VaR metrics.

\subsection{Multi-Objective Optimization}\label{sec:moo-background}
Without loss of generality, we assume that MOO concerns the joint maximization of the $|K|$ objectives $f_1, ..., f_{|K|}$, where $f_k$ which needs to be minimized can be converted to a maximization problem via its negative. Broadly speaking, there does not typically exist a feasible solution that marginally optimizes each objective function simultaneously. Therefore, MOO approaches generally focus on Pareto-optimal solutions. A feasible solution is considered Pareto-optimal if no objective can improve without degrading another; in other words, if the solution is not Pareto-dominated by any other solution. The Pareto frontier is defined as the set of all Pareto-optimal solutions.

A common approach to MOO is to convert the $|K|$-dimensional DVH objective to a scalar in order to utilize standard optimization methods via a scalarization function. Scalarization functions typically take the form $s_\lambda:\mathbb{R}^{|K|}\rightarrow \mathbb{R}$, parameterized by $\pmb{\lambda}$ from some set $\Lambda$ in $L$-dimensional space~\citep{roijers_survey_2013, paria_flexible_2019}. 

The parameters $\pmb{\lambda} \in \Lambda$ can be viewed as weights, or relative preferences (which are often hidden and need to be elicited), on the objective functions in the scalarized optimization objective $\max_{x\in X} s_{\pmb{\lambda}}\big([f_1(x), \dots, f_{|K|}(x)]\big)$ for input space $X$. Then, the advantage of using scalarization functions is that the solution to maximizing $s_{\pmb{\lambda}}\big([f_1(x), \dots, f_{|K|}(x)]\big)$, for a fixed value of $\pmb{\lambda}$, may lead to a solution along the Pareto frontier of DVH metrics. 

A typical approach in MOO literature for comparing different sets of solutions on the Pareto frontier is the hypervolume indicator, which quantifies the volume of the dominated portion of the objective space. For some set of solutions $M \subseteq \mathbb{R}_+^{|K|}$, the hypervolume indicator can be defined as $\text{Vol}(\cup_{(y_1, ..., y_{|K|}\in M)} [0, y_1] \bigtimes ... \bigtimes [0, y_{|K|}])$, where $\text{Vol}()$ corresponds to the typical Lebesgue measure. Without loss of generality, the hypervolume indicator definition assumes a reference point of $0^{|K|}$. We leverage the hypervolume indicator as a metric in Section \ref{sec:eval-iterative-exploration}.

\subsection{Iterative Clinician Exploration}\label{app:iterative-clinician-exploration}

\subsubsection{Interactively Learning From Aim And Limit Values}\label{aim-limit-interaction-section}

The clinician interaction with the aim and limit values serves as a useful feedback mechanism, enabling the system to learn the clinician's hidden preferences ($\pmb{\lambda}^*$) and refine the displayed set of Pareto-optimal treatment plans ($M$). This learning process is inspired by the feedback interpretation and probabilistic modeling detailed in \citet{chen_multi-objective_2025}.

\noindent
\textbf{Feedback Interpretation.}
When a clinician adjusts an aim or limit slider, this action implicitly signals a shift in their region of interest within the DVH tradeoff space. The core assumption is that the clinician is now most interested in, or wishes to understand the implications of, treatment plans that are characterized by this new bound configuration. For instance, if a clinician tightens the limit for bladder dose (e.g., requiring Bladder$_{D2cc}$ to be lower), it implies a heightened concern for bladder toxicity, and plans that adhere to or are near this new stricter limit become highly relevant for evaluation. Conversely, relaxing a limit for PTV coverage (e.g., accepting a slightly lower PTV$_{V700}$) might indicate a willingness to explore tradeoffs that are now newly feasible under the relaxed PTV coverage. Adjustments to aim values similarly refine the clinician's aspirational targets.

\begin{figure}
    \centering
    \includegraphics[width=0.49\textwidth]{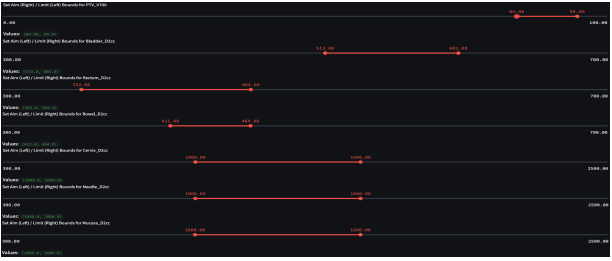}
    \caption{Aim and Limit Sliders for ALMo. Aim and limit sliders are shown for the planning target volume (PTV), all of the organs-at-risk (bladder, rectum, and bowel), and the hot spot regions (cervical, needle, and vaginal mucosa). These sliders are displayed at the top of the tool GUI.}
    \label{fig:sliders}
\end{figure}

Unlike the framework in \citet{chen_multi-objective_2025}, where a single bound modification was typically assumed per iteration, ALMo's interface (Figure~\ref{fig:sliders}) allows clinicians to adjust multiple aim and limit values for various DVH metrics simultaneously before requesting a new set of treatment plans. Each individual adjustment made by the clinician (e.g., changing the PTV aim, the Bladder limit, etc.) is interpreted as inducing a distinct preference ranking over the currently displayed set of treatment plans $M_{curr}$. The rationale for how each type of individual aim or limit adjustment translates into a ranking is adapted from \citet{chen_multi-objective_2025}:

\begin{itemize}
    \item \textbf{Limit Value Made More Restrictive} (e.g., increasing the minimum PTV$_{V700}$ limit or decreasing the maximum Bladder$_{D2cc}$ limit): When a clinician makes a limit value more stringent, they are refining the criteria for acceptable plans. We interpret this as a strong signal that their interest lies in treatment plans that satisfy this new, more demanding requirement. The most informative plans in $M_{curr}$ are those that now lie near, or just satisfy, this new limit. These plans are thus ranked higher. The critical signal is the clinician's directive to meet this new stricter constraint, and plans near this boundary are key to understanding achievable tradeoffs under this condition.

    \item \textbf{Limit Value Relaxed} (e.g., decreasing the minimum PTV$_{V700}$ limit or increasing the maximum Bladder$_{D2cc}$ limit): Conversely, when a clinician relaxes a limit value, they are expanding the set of acceptable plans. This action signals an interest in exploring solutions within this newly accessible part of the tradeoff space, perhaps because previous limits were too restrictive and prevented consideration of otherwise desirable plans (e.g., plans with excellent OAR sparing but slightly below the previous PTV coverage limit). The critical signal is the clinician's willingness to explore what was previously deemed unacceptable. Plans in $M_{curr}$ that fall within or are close to this newly opened region, particularly those that benefit from this relaxation (e.g., showing significant improvement in another metric), are ranked more favorably.

    \item \textbf{Aim Value Adjusted} (e.g., changing the PTV$_{V700}$ aim or the Rectum$_{D2cc}$ aim): An adjustment to an aim value represents a direct refinement of the clinician's aspirational target for a specific DVH metric. We interpret this as an explicit indication that they are now most interested in treatment plans whose value for that DVH metric is as close as possible to the new aim value. The critical signal is precisely this new target. Consequently, plans in $M_{curr}$ are ranked directly by their proximity (e.g., absolute difference) to the adjusted aim value in the modified DVH metric—the closer a plan's metric is to the new aim, the higher its rank.
\end{itemize}
If no change is made to a particular aim or limit value after reviewing $M_{curr}$, it is interpreted as an implicit confirmation of the current preference associated with that DVH metric's target or boundary. 

Each of these individually interpreted adjustments contributes to the overall likelihood model for updating the clinician's preference vector $\pmb{\lambda}$, as detailed in the subsequent section.

\noindent
\textbf{Modeling Aim-Limit Preferences.}
Given that multiple aim-limit values can be adjusted simultaneously in a single feedback step, we derive a set of $N_A$ implicit rankings, $\{\pi_j\}_{j=1}^{N_A}$, one for each of the $N_A$ individual bound adjustments made by the clinician. Each ranking $\pi_j$ is over the currently displayed plans $M_{curr}$.

To model the likelihood of observing this composite feedback, we employ a mixture of Plackett-Luce likelihoods \citep{zhao_learning_2016}. The likelihood of the set of DM-induced rankings from feedback step $curr$, given a preference vector $\pmb{\lambda}$ and the new set of bounds $\pmb{\alpha}_{curr+1}$, is:
\begin{equation}
\label{eq:almo_likelihood_mixture}
\begin{split}
    L(\text{feedback}_{curr} ; \pmb{\lambda}, M_{curr}, \pmb{\alpha}_{curr}, \pmb{\alpha}_{curr+1}) \\
    = \sum_{j=1}^{N_A} w_j \cdot p_{PL}(\pi_j ; \pmb{\lambda}, M_{curr}, {\pmb{\alpha}_{curr+1}})
\end{split}
\end{equation}
where $p_{PL}(\pi_j ; \pmb{\lambda}, M_{curr}, {\pmb{\alpha}_{curr+1}})$ is the standard Plackett-Luce likelihood for the $j$-th ranking $\pi_j$ given $\pmb{\lambda}$ and the SHFs ${\pmb{\alpha}_{curr+1}}$ defined by the newly set aim-limit values:
\begin{equation}
\label{eq:plackett_luce}
\begin{split}
    & p_{PL}(\pi_j ; \pmb{\lambda}, M_{curr}, {\pmb{\alpha}_{curr+1}}) \\
    & \quad = \prod_{h=1}^{\varrho} \frac{\exp\Bigl( s_{\pmb{\lambda}}( u_{f}\bigl(g(r_{\pi_j(h)})\bigr)) \Bigr)}{ \sum_{i=h}^{\varrho} \exp\Bigl( s_{\pmb{\lambda}} (u_{f}\bigl(g(r_{\pi_j(i)})\bigr)) \Bigr)}
\end{split}
\end{equation}
In Equation \ref{eq:plackett_luce}, the usage of ${\pmb{\alpha}_{curr+1}}$ is implicit in $u_f$ and $\pi_j = \bigl[\pi_j(1), \dots, \pi_j(\varrho)\bigr]$ is the clinician-induced ranking of the $\varrho$ points in $M$. 

The weights $w_j$ (where $\sum w_j = 1$ and $w_j \ge 0$) in Equation \ref{eq:almo_likelihood_mixture} reflect the relative importance or "strength" of the signal from each individual bound adjustment. These weights are calculated based on the magnitude of the change in the utility value of the adjusted bound, normalized across all $N_A$ adjustments made in that iteration: $w_j = \Delta U_j / \sum_{k=1}^{N_A} \Delta U_k$, where $\Delta U_j$ is the absolute difference in the SHF utility corresponding to the $j$-th bound before and after the clinician's adjustment. This heuristic ensures that larger, more significant adjustments by the clinician contribute more heavily to the overall likelihood.

\noindent
\textbf{Updating Beliefs.}
At each iteration, after the clinician provides feedback (i.e., adjusts one or more aim-limit sliders, leading to new bound parameters defining $\pmb{\alpha}_{curr+1}$), we update our posterior belief over the DM's latent preference vector $\pmb{\lambda}$ using Bayes' rule. 

The posterior distribution at feedback step $curr$ is:
\begin{equation}
\label{eq:almo_lambda_bayes_update}
\begin{split}
    p_{curr}(\pmb{\lambda}) \propto p(\pmb{\lambda}) \\
    \cdot \prod_{curr} L(\text{feedback}_{curr} ; \pmb{\lambda}, M_{curr}, \pmb{\alpha}_{curr}, \pmb{\alpha}_{curr+1})
\end{split}
\end{equation}
where $p(\pmb{\lambda})$ is the prior belief over $\pmb{\lambda}$ (which is the initial uninformative prior at $curr=0$), and $L(\text{feedback}_{curr} ; \dots)$ is the mixture likelihood defined in Equation~\eqref{eq:almo_likelihood_mixture}. This iterative Bayesian update allows ALMo to flexibly incorporate complex, multi-faceted feedback from clinicians and progressively refine its understanding of their underlying preferences for navigating the treatment plan tradeoffs. In practice, we sample from this posterior to guide the selection of new treatment plans to present in the subsequent iteration, as detailed in Section~\ref{mc-modeling-section}.

\subsection{Experimental Setup}\label{sec:experiental-setup}
All components of ALMo were retrospectively tested on cervical cancer brachytherapy clinical cases. In total, we used 25 cases. Each case consisted of the following applicators: tandem, left ovoid, right ovoid, and, on average, three needles. For each case, a broad selection of treatment plans is optimized and compared against a treatment plan which was planned with a commercial treatment planning system, ARIA Oncology Information System from Varian Medical Systems, and deployed in the clinic. We used Python and Streamlit to build the overall tool user interface, GurobiPy \citep{gurobi} for the multi-objective treatment plan optimization linear program, and HiPlot \citep{hiplot} for the interactive parallel coordinates plot display. For our experiments, ALMo was run locally on a computer with a 2.4 Ghz 8-Core Intel Core i9 processor.

Prescription dosage was determined to be 700 cGY and hot spots are defined to be regions with 200\% of that amount of radiation dosage. As the set of structures, or objectives, we optimize over, K = \{PTV, Bladder, Rectum, Bowel, Cervical Region, Vaginal Mucosa Region, Needle Region\}, where the latter three are in the set $H$ and represent the structures which we aim to reduce hot spots for -- the cervical region, the vaginal mucosa region, and surrounding needle regions. We use the following values for the initial values of $(\alpha^{aim}, \alpha^{lim})$: $\text{PTV}_{V700}$: (0.95, 0.90), $\text{Bladder}_{D2cc}$: (513, 601), $\text{Rectum}_{D2cc}$: (352, 464), $\text{Bowel}_{D2cc}$: (411, 464), $\text{Cervical}_{D2cc}$: (800, 1600), $\text{Mucosa}_{D2cc}$: (800, 1600), $\text{Needle}_{D2cc}$: (800, 1600). Most of the aforementioned objectives, $\text{PTV}_{V700}$, $\text{Bladder}_{D2cc}$, $\text{Rectum}_{D2cc}$, and $\text{Bowel}_{D2cc}$, were based on nationally recognized recommendations from ABS. The remaining objectives and values were recommended based on internal quality criteria. The metrics reported were all calculated using structure voxel sizes equal to the CT image resolution. During plan optimization, the structures were subsampled to approximately 500 voxels each to reduce computational time. We did not find this to materially affect our numerical results.

\noindent
\textbf{Automated Planning Parameter Setup.} 
For the Reference Plan Initialization Algorithm, we used the following values for $\alpha^{aim}: \text{Bladder}_{D2cc}: \text{513 cGY}, \text{Rectum}_{D2cc}: \text{352 cGY}, \text{Bowel}_{D2cc} = \text{411 cGY}$, as described above. For our experiments, we used the following values of $R:$ $R_{cervical} = \{2.0, 2.35, 2.70\}$, $R_{mucosa} = \{2.0, 2.12, 2.24\}$, and $R_{needles} = \{2.0, 5.0, 8.0, 11.0\}$ to determine the right-hand side values of the voxel-wise dosage constraints, as described in Section \ref{ref-plan-init-section}. For the Limiting Organs Detection Algorithm, we used in our experiments the following values for $R_{cervical}$, $R_{mucosa}$, and $R_{needles}: \{0.4, 0.6, 0.8, 1.0, 1.2\}$, whereas $R$ = $\{2.5\}$ for $R_{PTV}$, $R_{Bladder}$, $R_{Rectum}$, and $R_{Bowel}$. For the Epsilon-Ratio Method we used in our experiments a step size of 0.01.

\noindent
\textbf{Treatment Plan Optimization.}
To create the artificial structures for the cervical and vaginal mucosa regions, we used all voxels within a distance of 1 voxel of the PTV structure contour. We then used the heuristic of selecting 18\% of the voxels in the artificial ring to be used as the vaginal mucosa region artificial structure. The remaining voxels were optimized as the cervical region artificial structure. For the surrounding needle regions, we used a distance of 2 voxels.

\noindent
\textbf{Iterative Clinician Exploration.}
To compute the coarse grid search $R^{\dagger}$ described in Algorithm \ref{algorithm:dvh-sampling}, we used $\Delta_k^\dagger = 0.05$ $\forall k \in O$ and $\Delta_k^\dagger = 0.30$ $\forall k \in H$. To compute the fine-grained, linearly interpolated space, $R^{\ddagger}$, we used $\Delta_k^\ddagger = 0.02$ $\forall k \in O$ and $\Delta_k^\ddagger = 0.10$ $\forall k \in H$. The Ray API \citep{moritz_ray_2018} was used to parallelize computations for computing $R^{\dagger}$ such that, on average, 500 instances of Equation (\ref{equation:moo}) could be computed within 110 seconds. 

At each iteration, we displayed to the clinicians $|M| \approx 12$ optimized treatment plans, defined by the clinician's aim and limit values, $\varrho^D = 20$ and $\varrho = 12$. If MoSH-Sparse (Algorithm \ref{algorithm:saturate}) returns a sparse set of treatment plans $|M| < 12$, we use a naive greedy algorithm to solve Equation (\ref{submodular-formulation2}) and select $12-|M|$ additional treatment plans from $M^D$ to display to the clinicians. We do this to avoid iterations where clinicians are shown only a few treatment plans, which may decrease confidence in the understanding of the dosage tradeoff surface. For the initial iteration, we used the aforementioned values for $(\alpha^{aim}, \alpha^{lim})$. For subsequent iterations, the decision-maker was free to adjust the aim and limit values as they wish. For the SHF, we used $\beta = 0.08$.

\end{document}